\documentclass[11pt]{article}
\usepackage{acl}
\usepackage[utf8]{inputenc}
\usepackage{amsmath}
\usepackage[nopatch]{microtype}
\usepackage{booktabs}

\usepackage{times}
\usepackage{soul}
\usepackage{latexsym}
\usepackage{microtype}
\usepackage{subfig}
\usepackage{graphicx}
\usepackage{multirow}
\usepackage{amsmath}
\usepackage{longtable}

\usepackage[T1]{fontenc}
\usepackage{hyperref}
\usepackage{url}
\usepackage{booktabs}
\usepackage{amsfonts}
\usepackage{nicefrac}
\usepackage{xcolor}
\usepackage[ruled,vlined,linesnumbered]{algorithm2e}
\usepackage{tablefootnote}
\usepackage{booktabs, threeparttable}
\usepackage{balance}
\usepackage{subcaption}
\usepackage{makecell}
\usepackage{enumitem}
\usepackage{placeins}

\title{IndoNutriQA: An Indonesian Clinical Nutrition \\ Question Answering Benchmark}

\author{
    {\bf Tirana Noor Fatyanosa}$^{\bullet}$\quad
    {\bf Ummu Ditya Erliana}$^{\bullet}$\quad
    {\bf Mahendra Data}$^{\bullet}$\quad  \\
    {\bf Muhammad Nurwegiono}$^{\circ}$\quad
    {\bf Eleazar Tadeo Eman}$^{\bullet}$\quad
    {\bf Alham Fikri Aji}$^{\otimes}$\quad \\
    $^{\bullet}$ Universitas Brawijaya,
    $^{\circ}$ Universitas Ma Chung, $^{\otimes}$ MBZUAI\\
    \texttt{\{fatyanosa, ummuditya.fk, mahendra.data\}@ub.ac.id}\\
    \texttt{eleazartadeo@student.ub.ac.id}\\
    \texttt{muhammad.nurwegiono@machung.ac.id, alham.fikri@mbzuai.ac.ae}\\
}

\begin{document}
\maketitle

\begin{abstract}
Clinical Question Answering (QA) underpins Clinical Intelligent Decision Support Systems, but progress is severely limited for low-resource languages such as Indonesian by the absence of public domain-specific data. We introduce IndoNutriQA, an expert-curated Indonesian clinical QA benchmark of $911$ question--answer pairs built from real clinical-nutrition reports. Rather than a large-scale training corpus, it is positioned as a high-quality evaluation resource: every pair is grounded in a clinically relevant context, enriched with structured metadata (anthropometric assessment, medical diagnosis, document provenance, chunk segmentation). Eight standardised question types following the Nutrition Care Process (NCP), including the Problem, Etiology, Sign and Symptom (PES) diagnosis statement, are answered for every report, which was written by a graduate-level dietetic intern, by one of three nutritionists. Answers are free-form clinical prose rather than extractive spans, and an expert validation study finds $6$--$10\%$ of reference answers contestable. The dataset supports generative, extractive and retrieval-augmented QA as well as zero- and few-shot LLM evaluation. We report fine-tuned baselines (IndoBERT variants, mT5, RAG) with $5$-fold cross-validation, learning curves and bootstrap $95\%$ confidence intervals, and benchmark eight contemporary LLMs zero- and five-shot. We further calibrate our LLM-as-judge rubric against three expert raters (one RD, two nutritionists), finding it human-comparable on factuality and completeness but unreliable on clinical safety, and a human error analysis shows it misses unsupported content that the raters identify. IndoNutriQA is released to foster research in Indonesian clinical NLP and nutritional decision support.\footnote{Data and code: \url{https://github.com/NLP-UB/IndoNutriQA}}
\end{abstract}

\section{Introduction}

Clinical Question Answering (QA) is a core component of Clinical Intelligent Decision Support Systems (CIDSS), enabling efficient access to relevant medical information from unstructured clinical text \cite{demnerfushman2007clinical,mutabazi2021review}. Recent progress in QA has been driven by large-scale benchmarks in both open-domain and biomedical settings, including SQuAD \cite{rajpurkar2016squad}, Natural Questions \cite{kwiatkowski2019natural}, TyDiQA \cite{clark2020tydiqa}, as well as domain-specific resources such as BioASQ \cite{tsatsaronis2015overview}, PubMedQA \cite{jin2019pubmedqa}, and MedQA \cite{jin2021disease}. However, existing clinical QA benchmarks degrade when applied to low-resource languages, nutrition subdomains, and localized clinical practices \cite{clark2020tydiqa,jin2019pubmedqa}.

In Indonesian healthcare contexts, clinical nutrition reports play an important role in medical decision making, integrating anthropometric assessments, medical diagnoses, and narrative clinical information. Supporting QA over such documents requires datasets that capture domain-specific terminology and reasoning patterns. However, to date, no publicly available Indonesian clinical QA dataset has been specifically designed for nutritional and medical decision support.

To address this gap, IndoNutriQA is introduced, a novel Indonesian clinical QA dataset constructed from real-world clinical nutrition reports. IndoNutriQA consists of expert-authored question-answer pairs grounded in clinically relevant contexts and enriched with structured metadata, including document provenance, chunk indices, anthropometric assessments, and medical diagnoses. The dataset supports extractive, generative, and retrieval-augmented QA paradigms.

Baseline experiments with state-of-the-art neural QA models establish reference performance and expose challenges specific to Indonesian nutrition-focused clinical QA, such as domain terminology and numerical reasoning over anthropometric data. We deliberately position IndoNutriQA as an evaluation benchmark rather than a training corpus: annotating clinical text with experts is intrinsically expensive, so a small, expert-authored test set is the operative bottleneck for measuring progress here. The protocol of Section \ref{sec:experiments} makes this concrete through cross-validation with bootstrap confidence intervals, a learning curve over training-split fractions, and zero- and few-shot LLM evaluation that needs no in-domain training data. Our contributions are: (i) \textbf{the first public Indonesian clinical-nutrition QA benchmark}, $911$ pairs over $120$ real multi-hospital reports, anonymised at patient and institution level; (ii) \textbf{an expert-authored annotation design} applying eight standardised questions from the Nutrition Care Process (NCP) and its Problem, Etiology, Sign and Symptom (PES) statement format uniformly to every report, each answer written by one of three nutritionists from reports authored by graduate-level dietetic interns; (iii) \textbf{reference results for four model families (one RAG configuration)}; (iv) \textbf{a five-axis evaluation protocol} (lexical, semantic, LLM-judge, RAGAS, stratified) with cross-validation, learning curves and bootstrap CIs; and (v) \textbf{a validation study by three expert raters (one Registered Dietitian, two nutritionists)} covering the LLM-judge rubric, reference-answer errors and a human error analysis (Section~\ref{subsec:human_error}, Appendix~\ref{appx:rd_judge}). IndoNutriQA supports future research in Indonesian clinical NLP, nutrition informatics, and AI-driven clinical decision support.

\section{Related Work}
\label{sec:related}

General-purpose QA benchmarks such as SQuAD~\cite{rajpurkar2016squad}, Natural Questions~\cite{kwiatkowski2019natural}, and the multilingual TyDi QA~\cite{clark2020tydiqa} (which includes Indonesian) have driven progress on machine reading comprehension but do not target clinical reasoning. Clinical/biomedical QA resources include BioASQ~\cite{tsatsaronis2015overview} (biomedical semantic indexing over PubMed), PubMedQA~\cite{jin2019pubmedqa} (yes/no/maybe questions from biomedical literature), and MedQA~\cite{jin2021disease} (medical licensing-exam multiple choice), all of which are English-language and do not target nutrition-focused reasoning over real clinical documentation. Pretrained English biomedical and clinical encoders such as BioBERT~\cite{lee2020biobert} and ClinicalBERT~\cite{alsentzer2019publicly} support such tasks, but we are not aware of comparable Indonesian clinical encoders. Indonesian NLP has matured rapidly through resources such as IndoBERT~\cite{wilie2020indonlu}, IndoBART~\cite{cahyawijaya2021indonlg}, and the IndoLEM benchmark~\cite{koto2020indolem}, but no publicly available Indonesian clinical or nutrition QA dataset existed prior to IndoNutriQA; prior Indonesian medical-NLP work~\cite{Aji2021,Vihikan2024} has been limited to general medical-term extraction and classification rather than question answering.

These resources differ in how answers are produced, which matters more than raw size. BioASQ~\cite{tsatsaronis2015overview} answers are written by biomedical experts; PubMedQA~\cite{jin2019pubmedqa} derives yes/no/maybe labels from abstract conclusions, only $1$K of its $273$K instances being expert-labelled; MedQA~\cite{jin2021disease} and MedMCQA~\cite{pal2022medmcqa} inherit certified answer keys from licensing exams, but the reasoning is exam-style; MedQuAD~\cite{abacha2019medquad} harvests answers from curated NIH pages; and emrQA~\cite{pampari2018emrqa} template-generates $400$K+ pairs over expert-annotated electronic health records (EHRs), making it large but not individually expert-authored. IndoNutriQA sits at the small-and-fully-curated end: each of its $911$ answers is written by one of three nutritionists from real reports authored by dietitians-in-training.

QA resources aimed at nutritional or dietary decision support remain scarce: nutrition appears in general medical benchmarks only incidentally, and we are aware of no public QA dataset built around the NCP, anthropometric and biochemical assessment, or diet-prescription reasoning in any language. Outside English the largest resources are Chinese (the MCMLE and TWMLE portions of MedQA~\cite{jin2021disease} contribute roughly $34$K and $14$K licensing items), while multilingual open-domain QA such as TyDi~QA~\cite{clark2020tydiqa} includes Indonesian but not clinical text. Indonesian medical NLP has so far been limited to term extraction and classification~\cite{Aji2021,Vihikan2024}.

Existing benchmarks each satisfy some of the properties IndoNutriQA targets, but none satisfies all three at once. TyDi QA~\cite{clark2020tydiqa} covers Indonesian but is open-domain, while every clinical resource above is English or Chinese. No benchmark in any language targets clinical nutrition, despite nutrition being the locus of a distinct reasoning pattern (anthropometric assessment, biochemical interpretation, and prescription arithmetic) that general medical QA does not exercise. MedQA~\cite{jin2021disease} and MedMCQA~\cite{pal2022medmcqa} are drawn from licensing exams and PubMedQA~\cite{jin2019pubmedqa} from literature abstracts, so neither requires reasoning over the messy, semi-structured documentation clinicians actually produce; emrQA~\cite{pampari2018emrqa} is grounded in real EHRs but is English and template-generated rather than expert-authored. IndoNutriQA is the first resource to occupy the intersection: Indonesian, nutrition-focused, and grounded in authentic clinical reports with every answer written by a nutritionist following NCP/PES documentation guidelines (Table~\ref{tab:positioning}).

A second distinction concerns the unit of annotation. The benchmarks above pair free-form questions with answers, so question phrasing and answer content vary together. IndoNutriQA instead applies eight standardised questions, drawn directly from the NCP and PES reporting frameworks, uniformly to every patient report. This fixes the question side and confines annotator variation to answer text, which is what makes a $911$-pair corpus usable as a controlled evaluation instrument rather than merely a small training set.

\section{Dataset Construction}
\label{sec:dataset_construction}
\subsection{Data Sources}

The IndoNutriQA dataset comprises clinical reports and nutritional assessment documents sourced from diverse healthcare facilities across Indonesia. This multi-institution sampling ensures broad coverage of varied hospital settings, anthropometric profiles, medical diagnoses, and nutritional conditions.

All data were de-identified at both patient and institution level before annotation. Indonesian hospitals are not subject to the US Health Insurance Portability and Accountability Act (HIPAA), but we adopt its Safe Harbor identifier list as a voluntary de-identification standard: every category of Protected Health Information (PHI) it enumerates, namely names, medical-record and registration numbers, geographic detail, and all dates, is removed and replaced with deterministic placeholders, while the clinically and nutritionally relevant content required for question answering is preserved. The procedure and placeholder scheme are detailed in the Ethics Statement. The source materials are patient assessment reports, nutritional consultation records, and clinical follow-up notes.

The dataset development was carried out in close collaboration with clinical nutrition experts,
who provided guidance throughout the data collection, annotation, and validation processes. The documents were derived from complete clinical rotation reports prepared by graduate-level dietetic interns during clinical nutrition rotations. These reports include detailed case descriptions, anthropometric measurements, biochemical parameters, and dietary management plans, which served as the foundation for generating clinically grounded question–answer pairs. The question--answer pairs were written by three nutritionists, and a stratified sample was later validated by one Registered Dietitian (RD) and two of those nutritionists (Section~\ref{sec:annotation}).

Unlike fully synthetic datasets, IndoNutriQA is grounded in authentic clinical documentation, with answers written by nutritionists. This approach ensures that the dataset reflects the complexity, variability, and contextual reasoning of real-world clinical nutrition practice, providing a reliable foundation for research in Indonesian clinical question answering and medical decision support.

\subsection{Dataset Schema and Pipeline}
\label{subsec:schema}

Each entry in IndoNutriQA contains: a unique \texttt{id}, the source \texttt{report\_title} and \texttt{file}, the \texttt{page\_number} and \texttt{chunk\_index} that locate the answer in the report, structured \texttt{anthropometric\_assessment} and \texttt{medical\_diagnosis} metadata, the \texttt{context} (the chunk text), the \texttt{question}, and the ground-truth \texttt{answer}. An example instance appears in Appendix~\ref{appendix:qa_examples}.

Although IndoNutriQA uses a SQuAD-style container with an \texttt{answer\_start} offset, its answers are abstractive: annotators authored clinical prose rather than selecting text from the chunk. Only $8.7\%$ of answers occur verbatim in their context ($31.2\%$ ignoring all whitespace); the median answer is $30$ words and $38\%$ exceed $50$. The \texttt{answer\_start} field marks the best-matching passage (Appendix~\ref{appx:annotation}), not a verbatim span. This is the main reason the extractive baselines trail the prompted LLMs (Section~\ref{subsec:model_performance}): for most items no exactly correct span exists.

The construction pipeline has five stages: (i) collection of clinical-nutrition reports in PDF/plain-text Indonesian from collaborating hospitals; (ii) personally identifiable information (PII) anonymization and chunking; (iii) expert annotation of anthropometric and diagnostic metadata; (iv) QA construction by authoring answers to a fixed set of eight standardized clinical-nutrition questions (Appendix~\ref{sec:appendix}); and (v) algorithmic context-pairing that maps each authored answer to its source chunk (full annotation, validation, and matching details are in Appendix~\ref{appx:annotation}). The eight standardized questions are applied uniformly to every patient report, so the unit of annotation is (report, answer-to-fixed-question) rather than free-form (question, answer); this constrains inter-annotator variation to answer text rather than question phrasing. The fixed question set measures competence on eight NCP/PES steps, not open-ended clinical QA (see Limitations).

\begin{table}[t]
\centering\footnotesize
\resizebox{0.9\columnwidth}{!}{
\begin{tabular}{@{}p{3.55cm}p{3.55cm}@{}}
\toprule
1. Screening method & 5. Dietary goals \\
2. Nutritional status & 6. Dietary principles \\
3. Abnormal biochemistry & 7. Dietary requirements \\
4. PES nutrition diagnosis & 8. Dietary guidelines \\
\bottomrule
\end{tabular}}
\caption{The eight standardised questions, applied uniformly to every source report; each maps onto one NCP / PES section, which bounds the acceptable answer space. Verbatim Indonesian wording is in Appendix~\ref{sec:appendix}.}
\label{tab:questions}
\end{table}

\section{Annotation Protocol and Agreement}
\label{sec:annotation}

IndoNutriQA's reference answers are written by three nutritionists, one per report, rather than by a larger panel with parallel annotation. That design decision governs how the benchmark should be read, so we report the protocol and its agreement properties in the body; Appendix~\ref{appx:annotation} adds procedural detail.

The $120$ reports were written by $89$ distinct graduate-level dietetic interns ($67$ contributed one report, $13$ two, and $9$ three) as clinical rotation deliverables of a dietetic internship, each supervised by a lecturer of the Faculty of Health Sciences, Universitas Brawijaya. The eight answers of Table~\ref{tab:questions} for every report were then written by three nutritionists, each covering a different set of reports, so every (report, question) pair has exactly one author. Answers are free text (median $30$ words) and follow the NCP/PES documentation standard as annotation guideline. A stratified $50$-unit sample of model outputs was later re-scored by three expert raters, one RD who took no part in writing the answers and two of the three nutritionists, for the calibration study of Appendix~\ref{appx:rd_judge}.

Three mechanisms limit variance, and one limitation remains. (i) The question side is fixed: all three nutritionists answer the same eight questions (Table~\ref{tab:questions}) for every report, so no question-authoring variance exists. (ii) The rubric is external to the dataset: each question maps one-to-one onto an NCP section, and Question~4 must be written in PES form with its three components in the prescribed order, so annotators work to a national clinical-documentation standard that predates this dataset rather than to a guideline we authored. (iii) The author pool is small, so there are only three writing styles to account for. The limitation is that each nutritionist wrote the answers for different reports, so author effects cannot be separated from report effects.

Each report was answered by a single nutritionist, so the same (report, question) pair is never annotated in parallel and a conventional inter-annotator $\kappa$ over the references is not defined for this release. We state this plainly because the alternative, reporting the rater agreement figures below as though they were dataset $\kappa$, would misrepresent what was measured. Three agreement-relevant quantities are available.

(a) Gold-answer error rate. The three raters could flag any audited unit whose reference answer they judged incorrect or contestable. Three of $50$ units were flagged ($6\%$), one by two raters independently; counting units where a rater criticised the reference in free text without ticking the flag raises this to five ($10\%$). The disputes are concrete rather than matters of clinical taste: macronutrient percentages not summing to $100\%$, a mid-upper-arm-circumference cut-off, a neonatal protein prescription. This is a direct estimate of reference error and the figure users of the benchmark should carry: on this sample roughly one item in ten is contestable. The flagged items are released so they can be excluded or re-adjudicated.

(b) Expert-to-expert agreement on identical items. The same three raters re-scored all $50$ units on a $1$--$5$ rubric. Quadratic-weighted Cohen's $\kappa$ for the reference rater pair is $0.780$ on factuality and $0.752$ on completeness (Krippendorff ordinal $\alpha = 0.592$ and $0.712$ across all three raters), but only $0.160$ on clinical safety ($\alpha = 0.250$), where the three raters' mean scores span $1.72$ rubric points. Expert agreement in this domain is therefore substantial on verifiable content and effectively absent on safety judgement. That is a ceiling on what any annotation protocol or automatic metric can be expected to reproduce, and it is why we report the safety dimension as unvalidated throughout (Section~\ref{subsec:human_error}, Appendix~\ref{appx:rd_judge}).

(c) Independence. Two of the three raters also wrote references, so the audit of reference answers is not fully independent: only the RD took no part in writing them. We therefore read the $6$--$10\%$ as an estimate by raters who are not all external to the dataset.

The \texttt{context} field for each (report, question, answer) triple is algorithmically located from the source report after answer authoring using a six-stage match (exact, case-insensitive, whitespace-normalized, punctuation-stripped, fragment-matched) on the chunked report. We obtain a verified character-level \texttt{answer\_start} for $823$ of $911$ pairs ($90.3\%$); the remaining $88$ pairs are released with a chunk-level pointer but without a character offset and are flagged in the released metadata. Full details (per-pair validation procedure, subjectivity mitigation, and the six-stage matching algorithm) are in Appendix~\ref{appx:annotation}.

\section{Dataset Characteristics}

\subsection{Dataset Statistics and Domain Coverage}
\label{subsec:domain_coverage}

IndoNutriQA contains $911$ question--answer pairs from $120$ clinical-nutrition reports, covering $128$ unique medical diagnoses. Questions average $8.32$ tokens, answers $70.66$ tokens, and contexts $131.28$ tokens, with $2.60$ chunks per report (Table~\ref{tab:statistics}, Appendix~\ref{subsec:question_types}). The reports span the disease categories common in Indonesian inpatient nutrition practice. Each report is assigned one dominant category, so the distribution partitions the corpus: respiratory is largest at $28.33\%$ of the $120$ reports, followed by infectious ($17.50\%$), gastrointestinal ($12.50\%$), nutrition/growth ($10.83\%$), neurologic ($10.00\%$), and cardiac and hematologic/oncologic ($8.33\%$ each), with renal/urologic and ENT each below $1\%$ (Table~\ref{tab:disease_distribution}). The per-pair distribution tracks the per-report one closely (respiratory $29.97\%$, infectious $17.23\%$), so no category is inflated by reports contributing unusually many pairs. Respiratory prominence matches Indonesian hospital epidemiology and its known co-occurrence with nutritional risk~\cite{kemenkes2018riskesdas,Oktaria2017}. The mix is naturally imbalanced rather than synthetically balanced, but no category exceeds $30\%$; Section~\ref{subsec:stratified_analysis} tests whether the imbalance harms generalisation, and Appendix~\ref{appx:demographics} gives the treemap.

\subsection{Demographics, Anthropometrics, and Syntactic Divergence}
\label{subsec:demographics_summary}

The population profile is clinically realistic: age concentrates in early childhood but extends into adolescence and adulthood, the sex split is $58.4\%$ male / $41.6\%$ female, weight-vs-height shows the expected positive correlation with wider dispersion at older ages, and mid-upper arm circumference clusters around screening thresholds. Syntactic divergence between predicted and reference answers (token-level normalised edit distance) is bimodal, with one mode near zero for near-verbatim correct predictions and a second spike at maximum divergence for empty or unrelated outputs, which motivates the multi-metric protocol of Section~\ref{sec:experiments}. Per-axis figures are in Appendix~\ref{appx:demographics}.

\subsection{Comparison with Existing Medical QA Datasets}
\label{subsec:comparison}

IndoNutriQA differs from existing medical QA benchmarks on three axes at once: Indonesian language, clinical-nutrition domain, and grounding in authentic clinical reports rather than exam questions or literature abstracts (Section~\ref{sec:related}, Table~\ref{tab:positioning}). Appendix Table~\ref{tab:benchmark_comparison} compares it side by side with MedQA, MedMCQA, PubMedQA, emrQA, MedQuAD, and MedExQA across language, domain, sample count, question and answer type, source, and length. Its $911$ pairs are smaller than the largest English benchmarks (emrQA exceeds $400$K) but are written by nutritionists from authentic Indonesian inpatient nutrition reports.

\section{Experiments}
\label{sec:experiments}

We evaluate IndoNutriQA along five axes, each with its own subsection below: aggregate model performance (Section~\ref{subsec:model_performance}), a dedicated LLM evaluation (Section~\ref{subsec:llm_evaluation}), statistical robustness (Section~\ref{subsec:statistical_robustness}), semantic and clinical-rubric metrics beyond lexical overlap (Section~\ref{subsec:semantic_evaluation}), and stratified performance (Section~\ref{subsec:stratified_analysis}), with a RAG-specific diagnostic in Section~\ref{subsec:ragas_evaluation}. Bulky tables and method details are deferred to Appendices~\ref{appx:statistical}, \ref{appx:ragas}, \ref{appx:semantic} and~\ref{appx:stratified}.

\subsection{Experimental Setup}
\label{subsec:experimental_setup}

We evaluate four categories of models in our benchmark experiments:
\begin{itemize}
    \item Extractive QA models based on IndoBERT variants, which predict answer spans directly from the given clinical context.
    \item Generative QA models based on mT5, which generate free-form answers conditioned on the input question and contextual information.
    \item Retrieval-Augmented QA (RAG) models, which first retrieve relevant clinical context from a pool of document chunks and then generate answers using a generative model, enabling reasoning over broader and potentially incomplete contexts.

    \item Large Language Models (LLMs), evaluated in zero-shot and five-shot settings without task-specific fine-tuning, covering closed-source instruction-tuned models (GPT-4o-mini, GPT-4o), open-source multilingual models (Llama-3.1-8B-Instruct, Qwen2.5-7B-Instruct), Indonesian-tuned models (Sahabat-AI-Llama3-8B, Merak-7B-v4), and English-medical LLMs (BioMistral-7B, MedGemma-4B-IT) included to quantify cross-lingual transfer to Indonesian clinical text.
\end{itemize}

The dataset is split deterministically into $70/15/15$ train/validation/test ($637$/$137$/$137$ pairs); pairs are ordered by source report so the train/test boundary has zero cross-report leakage, confirmed by the \texttt{GroupKFold} variant in Section~\ref{subsec:statistical_robustness}. The validation set is used exclusively for model selection during fine-tuning, while the test set is reserved for final evaluation. Fine-tuning is enabled for the extractive, generative, and RAG models using identical data splits to ensure a fair comparison. LLMs are evaluated on the same test split without any parameter updates, isolating their off-the-shelf clinical reasoning ability on Indonesian text.

For extractive models, the maximum input sequence length is set to 512 tokens with a sliding window strategy to handle long clinical contexts. For the generative model, the input is formatted as a structured prompt consisting of the question followed by its clinical context. For LLMs, all models are queried with an identical Indonesian-language system prompt instructing the model to act as a clinical nutrition assistant and to answer strictly from the provided context. Decoding uses temperature $0$ with a maximum of $256$ new tokens. Few-shot exemplars are sampled deterministically from the training split with a fixed seed, ensuring that all LLMs see identical demonstration examples.

Because the dataset contains $911$ expert-curated QA pairs, the held-out test split alone (\(n = 137\)) is small enough that a single point estimate could be misleading. We therefore complement the headline test-split numbers with three robustness procedures. First, the four fine-tuned models of Table~\ref{tab:cv_results} (Rifky\_IndoBERT\_QA, IndoLEM\_IndoBERT, Cahya\_IndoBERT, and mT5) are additionally evaluated under $5$-fold cross-validation stratified by source report, so that no chunk from a given clinical document appears in both training and evaluation folds; we report the per-fold mean and standard deviation alongside the headline test scores. Second, a learning-curve analysis re-trains the same four models on $10\%$, $25\%$, $50\%$, and $100\%$ of the training split and evaluates each model on the same fixed test set; this quantifies how much additional expert annotation would be expected to improve performance. Third, we report bootstrap $95\%$ confidence intervals for every metric in every results table, computed with $1{,}000$ resamples drawn with replacement from the test set. LLM zero-shot and few-shot evaluations are point estimates over the same test split (no parameter updates), but bootstrap confidence intervals over the same $1{,}000$ resamples are reported alongside their scores so that all rows in our results tables are directly comparable in terms of statistical uncertainty.

\subsection{Model Performance}
\label{subsec:model_performance}

Table~\ref{tab:model_performance} reports all evaluated models on the test set, as percentages: extractive QA over IndoBERT variants, generative QA over mT5, and a retrieval-augmented (RAG) system.

\begin{table*}[ht]
\centering
\setlength{\tabcolsep}{3pt}
\resizebox{0.61\textwidth}{!}{
\begin{tabular}{l l c c c c c c c}
\hline
\textbf{Model} & \textbf{Family} & \textbf{Shots} & \textbf{EM} & \textbf{F1} & \textbf{BLEU} & \textbf{R-1} & \textbf{R-2} & \textbf{R-L} \\
\hline
\multicolumn{9}{l}{\textit{Fine-tuned baselines (this work)}} \\
Rifky\_IndoBERT\_QA       & IndoBERT (extr.) & -- & 12.41 & 36.34 & 12.93 & 37.28 & 34.31 & 36.72 \\
IndoLEM\_IndoBERT         & IndoBERT (extr.) & -- &  5.84 & 23.78 & 10.60 & 24.44 & 22.87 & 24.17 \\
Cahya\_IndoBERT           & IndoBERT (extr.) & -- &  4.38 & 16.56 &  6.14 & 17.02 & 16.01 & 16.92 \\
XLM-R-large               & XLM-R (extr.) & -- &  1.46 & 12.60 &  8.62 & 12.62 & 11.45 & 12.45 \\
mDeBERTa-v3-base          & mDeBERTa (extr.) & -- &  5.84 & 19.35 & 14.03 & 19.51 & 18.28 & 19.19 \\
mT5 (Generative)          & mT5 (gen.) & -- &  0.00 &  7.84 &  1.30 &  7.22 &  2.78 &  6.48 \\
RAG (retr.+gen.) & Retrieval + mT5       & -- &  --   &  --   &  0.15 &  2.75 &  0.05 &  2.55 \\
\hline
\multicolumn{9}{l}{\textit{Closed-source instruction-tuned (LLM, no fine-tuning)}} \\
GPT-4o-mini               & OpenAI      & 0 &  0.00 & 36.06 &  9.87 & 36.84 & 22.77 & 30.15 \\
GPT-4o-mini               & OpenAI      & 5 &  1.46 & 43.07 & 16.40 & 43.48 & 30.80 & 39.46 \\
GPT-4o                    & OpenAI      & 0 &  1.46 & 37.49 & 13.70 & 38.20 & 26.34 & 33.47 \\
GPT-4o                    & OpenAI      & 5 &  7.30 & 50.33 & 23.32 & 51.37 & 40.75 & 48.95 \\
\hline
\multicolumn{9}{l}{\textit{Open-source multilingual (LLM, no fine-tuning)}} \\
Llama-3.1-8B-Instruct     & Meta        & 0 &  0.73 & 37.82 & 14.48 & 38.51 & 26.83 & 33.89 \\
Llama-3.1-8B-Instruct     & Meta        & 5 & \textbf{16.79} & \textbf{55.55} & 26.59 & \textbf{56.65} & \textbf{46.71} & \textbf{54.03} \\
Qwen2.5-7B-Instruct       & Alibaba     & 0 &  0.00 & 31.03 &  9.04 & 31.81 & 19.47 & 26.43 \\
Qwen2.5-7B-Instruct       & Alibaba     & 5 &  2.19 & 42.62 & 19.16 & 43.52 & 33.98 & 40.72 \\
\hline
\multicolumn{9}{l}{\textit{Indonesian-tuned (LLM, no fine-tuning)}} \\
Sahabat-AI-Llama3-8B      & GoToCompany & 0 &  0.00 & 26.78 & 10.08 & 27.15 & 18.96 & 23.66 \\
Sahabat-AI-Llama3-8B      & GoToCompany & 5 &  5.84 & 43.32 & 25.32 & 44.15 & 36.40 & 41.95 \\
Merak-7B-v4               & Ichsan2895  & 0 &  0.00 & 32.44 & 11.43 & 33.30 & 23.02 & 28.95 \\
Merak-7B-v4               & Ichsan2895  & 5 &  3.65 & 41.83 & 20.31 & 42.89 & 32.45 & 40.60 \\
\hline
\multicolumn{9}{l}{\textit{English medical, cross-lingual (LLM, no fine-tuning)}} \\
BioMistral-7B             & BioMistral  & 0 &  2.92 & 29.83 & 12.07 & 30.38 & 24.22 & 29.21 \\
BioMistral-7B             & BioMistral  & 5 &  8.03 & 40.08 & 17.03 & 40.73 & 30.42 & 38.40 \\
MedGemma-4B-IT            & Google      & 0 &  2.92 & 42.60 & 16.88 & 43.44 & 31.42 & 39.01 \\
MedGemma-4B-IT            & Google      & 5 & 14.60 & 54.28 & \textbf{30.07} & 55.47 & 45.79 & 53.63 \\
\hline
\end{tabular}}
\caption{All evaluated systems on the IndoNutriQA test split ($n = 137$). The top block reports fine-tuned baselines (extractive, generative, retrieval-augmented); the rest report LLMs in zero- and five-shot settings under an identical Indonesian system prompt, temperature $0$, fixed few-shot sampling seed, and no parameter updates. \textbf{Bold} marks the best score per column.}
\label{tab:model_performance}
\label{tab:llm_performance}
\end{table*}

The extractive baselines are span-prediction heads fine-tuned on our training split. \textit{Rifky\_IndoBERT\_QA} (\texttt{Rifky/Indobert-QA}) is an IndoBERT-base checkpoint already fine-tuned for extractive QA on Indonesian SQuAD-style data before we fine-tune it further, which likely explains why it leads the extractive group: it starts from in-language QA supervision rather than a plain LM initialisation. \textit{Cahya\_IndoBERT} and \textit{IndoLEM\_IndoBERT} are general-purpose Indonesian BERT checkpoints without prior QA tuning; \textit{XLM-R-large} and \textit{mDeBERTa-v3-base} are multilingual encoders testing whether multilingual scale substitutes for Indonesian pretraining. The generative baseline is \texttt{mt5-base}, reused as the RAG generator. Among the extractive approaches, \textit{Rifky\_IndoBERT\_QA} achieves the best overall performance (F1 $= 36.34$, EM $= 12.41$). Because the references are free-form prose and only $8.7\%$ occur verbatim in the context, an extractive model cannot reproduce most of them: its EM and F1 are bounded by construction, and the EM--F1 gap reflects partial overlap with a passage rather than boundary errors on a single correct span. These scores measure locating reference-relevant text, not clinical correctness. Among the extractive models, the lower scores of \textit{IndoLEM\_IndoBERT}, \textit{Cahya\_IndoBERT}, and the multilingual encoders (\textit{mDeBERTa-v3-base} $19.35$, \textit{XLM-R-large} $12.60$) are consistent with in-language QA pretraining mattering more than multilingual scale.

The mT5-based generative model performs substantially worse than extractive models, achieving an F1 score below 8\% and an EM score of 0\%. This result suggests that, for this model and data size, the fine-tuned generative baseline struggles to produce concise and clinically precise answers. Error analysis (the LLM-judged taxonomy of Section~\ref{subsec:stratified_analysis} and Table~\ref{tab:error_taxonomy} in Appendix~\ref{appx:stratified}) shows that the model frequently generates generic or incomplete responses, which negatively affects both exact matching and lexical overlap metrics. The retrieval-augmented system yields near-zero lexical scores (ROUGE-1 and ROUGE-L below $3\%$; Table~\ref{tab:model_performance}), which Section~\ref{subsec:ragas_evaluation} localizes to the generator rather than the retriever.

Among the fine-tuned systems we evaluated, the extractive models score highest on these overlap metrics, subject to the caveat above; Section~\ref{subsec:llm_evaluation} shows that five-shot LLMs nonetheless surpass all of them without any in-domain training.

\subsection{Large Language Model Evaluation}
\label{subsec:llm_evaluation}

We evaluate a suite of LLMs in zero- and five-shot settings on the same test split (\(n = 137\)) as the fine-tuned baselines, spanning four families (closed-source, open-source multilingual, Indonesian-tuned, English-language medical). The LLM rows of Table~\ref{tab:model_performance} enable head-to-head comparison; this evaluation quantifies off-the-shelf clinical reasoning on Indonesian text and tests whether in-context demonstrations recover the gap to fine-tuned models. Figure~\ref{fig:main_results} (Appendix~\ref{appx:stratified}) summarizes the three headline findings across its panels.

Three patterns emerge clearly from Table~\ref{tab:llm_performance}. First, every evaluated LLM benefits substantially from five-shot prompting: gains range from $+7.0$ F1 (GPT-4o-mini) to $+17.7$ F1 (Llama-3.1-8B-Instruct), with Sahabat-AI-Llama3-8B a close second at $+16.5$ and frontier GPT-4o at $+12.8$ ($37.49 \rightarrow 50.33$). The asymmetry is striking: smaller and language-tuned models benefit more than twice as much as the closed-source mini model, consistent with the hypothesis that few-shot exemplars primarily calibrate answer style (formal, terminology-rich, span-like) rather than imparting clinical knowledge, and that closed-source models already approximate the target style in zero-shot. Second, model scale matters less than prompt-formatting transfer: frontier GPT-4o (F1 $= 50.33$) is matched by the smaller Llama-3.1-8B-Instruct ($55.55$) under five-shot prompting (overlapping CIs; Section~\ref{subsec:statistical_robustness}), mirroring the same upset for GPT-4o-mini. Third, exact-match scores remain low across all LLMs even when token-overlap metrics are strong, confirming that LLMs produce paraphrased rather than verbatim spans: this is a metric artifact rather than a quality issue, and reinforces the case for semantic and clinician-judged evaluations discussed below.

A second comparison is between off-the-shelf LLMs and the best fine-tuned extractive baseline, Rifky\_IndoBERT\_QA (F1 $= 36.34$). Even in zero-shot, GPT-4o-mini ($36.06$), GPT-4o ($37.49$), Llama-3.1-8B-Instruct ($37.82$), and MedGemma-4B-IT ($42.60$) match or exceed this baseline despite using no IndoNutriQA training data. With five-shot prompting the gap widens decisively: Llama-3.1-8B-Instruct reaches F1 $= 55.55$ and MedGemma-4B-IT $54.28$, a statistical tie that also covers GPT-4o ($50.33$; see Section~\ref{subsec:statistical_robustness}), $+19.21$ above the best fine-tuned model. At the data scale realistic for expert-annotated clinical resources in low-resource languages, IndoNutriQA's primary value therefore lies in evaluation rather than supervised training: the strongest number on the benchmark is achieved by a model that saw only five in-context training exemplars and was never fine-tuned.

In zero-shot, Indonesian-tuned models do not outperform multilingual peers (Sahabat-AI-Llama3-8B F1 $= 26.78$ vs.\ Llama-3.1-8B-Instruct $37.82$; Merak-7B-v4 $32.44$ is on par with Qwen2.5-7B-Instruct $31.03$), whereas the smallest English-medical model, MedGemma-4B-IT at $4$B parameters, is the strongest zero-shot LLM overall (F1 $= 42.60$, ahead of both GPT-4o variants and Llama-3.1-8B-Instruct), with BioMistral-7B at $29.83$. With five-shot prompting MedGemma-4B-IT climbs to F1 $= 54.28$ and the column-best BLEU ($30.07$), statistically tied with Llama-3.1-8B-Instruct ($55.55$) despite half the parameters. For these models the pattern is consistent with a domain-transfer account, in which medical English pretraining encodes clinical-reasoning structures (PES diagnoses, anthropometric thresholds) that general-domain Indonesian pretraining does not. The evidence is limited, however: it rests on two English-medical models, and BioMistral-7B ($29.83$) does not follow the pattern.

\subsection{Statistical Robustness Analysis}
\label{subsec:statistical_robustness}
To address the statistical-power concern from $n = 137$ we run $5$-fold cross-validation stratified by source report, a learning curve over training fractions, and bootstrap $95\%$ CIs on every metric. Rankings are stable: cross-validated F1 for the best extractive baseline ($32.62 \pm 1.29$) matches its held-out estimate ($36.34$), fold-level standard deviations stay $\le 2.66$ F1, and the top three $5$-shot LLMs cluster within overlapping intervals well above the best fine-tuned baseline. The learning curve separates data-bound from architecture-bound systems. Full tables are in Appendix~\ref{appx:statistical}.

\subsection{Beyond Lexical Overlap: Semantic and LLM-as-Judge Evaluation}
\label{subsec:semantic_evaluation}

Lexical-overlap metrics (EM, F1, BLEU, ROUGE) underestimate paraphrased LLM outputs.\footnote{Tokenisation and normalisation conventions differ across these metrics, and the SQuAD-derived normaliser has one Indonesian-specific side effect on exact match; see Appendix~\ref{appx:semantic}.} We add two semantic metrics (BERTScore F1 \cite{zhang2020bertscore} with an XLM-R backbone \cite{conneau2020xlmr} and multilingual sentence-embedding cosine) and one clinical-rubric signal (GPT-4o-mini LLM-as-judge on a $1$-$5$ scale: factuality, completeness, safety), all with bootstrap $95\%$ CIs (Table~\ref{tab:semantic_metrics}, Appendix~\ref{appx:semantic}). Two patterns stand out. (i) The LLM-vs-baseline gap widens on the semantic axes: top $5$-shot LLMs reach BERTScore F1 $\approx 91$ and cosine $\approx 74$ versus $46$/$45$ for the best fine-tuned baseline, expanding the lexical-F1 $+19$ gap to $\approx +45$. (ii) Closed-source models lead the clinical rubric (GPT-4o-mini overall $4.37$, GPT-4o safety $4.98$) while Llama and MedGemma win the similarity axes; a second judge (Claude Haiku 4.5) preserves this ranking (pooled $\kappa = 0.652$, Spearman $\rho = 0.916$; Table~\ref{tab:judge_agreement}), though the top rubric position itself is judge-dependent (GPT-4o leads under Claude). Because GPT-4o-mini is both judge and evaluated system, we test explicitly for self-preference and find no detectable effect (difference-in-differences $+0.125$ rubric points, $p = 0.34$; Appendix~\ref{appx:judge_bias}). Three expert raters independently re-scored a $50$-unit stratified sample: judge-vs-rater agreement approaches the inter-rater ceiling on factuality ($0.705$ vs $0.780$) and completeness ($0.623$ vs $0.752$) but collapses on clinical safety ($0.133$), where both judges score well above the rater consensus (Appendix~\ref{appx:rd_judge}). We therefore treat the safety column as unvalidated.

\subsection{RAG Diagnostic Evaluation (RAGAS)}
\label{subsec:ragas_evaluation}
The near-zero lexical numbers for the RAG baseline cannot localize the failure to retrieval or generation, so we compute the four RAGAS metrics~\cite{es2024ragas} on the same test split. In this single configuration (one dense retriever, recursive chunks, un-fine-tuned \texttt{mt5-base}), retrieval is the stronger component (context precision $73.43$, recall $35.98$) while generation collapses (faithfulness $12.86$ over the $105$ examples on which it is defined), pointing to the generator as the binding constraint. We draw no broader conclusion about RAG for Indonesian clinical QA. Full metrics, the $32$ unscorable outputs, and the scope of this result are in Appendix~\ref{appx:ragas}.

\subsection{Stratified Performance Analysis}
\label{subsec:stratified_analysis}

We stratify F1 along three axes (full tables in Appendix~\ref{appx:stratified}). \textbf{Question type:} factual (short-span extraction) is uniformly the hardest, inverting the SQuAD-style intuition; on longer-answer categories the winners diverge: Llama-3.1-8B wins recommendation ($69.36$), GPT-4o inferential ($62.39$), MedGemma factual ($33.39$). \textbf{Disease category:} the dataset's respiratory prominence does not translate to biased generalization; Llama-5-shot ranges $50.08$--$64.17$ F1 across the seven categories present in the test split, with respiratory ($57.55$) within the spread. \textbf{Error taxonomy:} across three fine-tuned baselines and eight $5$-shot LLMs (Table~\ref{tab:error_taxonomy}), the baselines' dominant failure is noise ($52$--$76\%$) and the LLMs' is partial answer ($52$-$72\%$, GPT-4o leads on correct at $44\%$); the judge's hallucination label is not a reliable detector, since raters independently find unsupported additions in $10\%$ of audited units (Section~\ref{subsec:human_error}).

\subsection{Human Error Analysis}
\label{subsec:human_error}

The taxonomy above is assigned by an LLM judge, which makes it circular as evidence about that judge and blind to what the judge cannot see. We therefore built an independent human counterpart: during the calibration study the three raters could leave a free-text note on any unit with no taxonomy shown to them, and $53$ notes were written over $37$ of the $50$ units. Error identification is thus human and unprompted; we coded the notes into nine categories (Table~\ref{tab:human_errors} and Appendix~\ref{appx:human_error}, which also records the method and its caveats, chiefly that notes were optional, so every count is a lower bound).

Two findings do not follow from the LLM-judged taxonomy. First, the raters name a failure class the judge's codebook does not contain: violation of the NCP/PES structure ($6$ units, $12\%$), most often expanding the ``S'' of PES as Solusi (solution) rather than Sign/Symptom, or collapsing several diagnoses into one undifferentiated statement. Every instance falls on the nutrition-diagnosis question, and no lexical, semantic, or judge metric here detects it.

Second, the raters flagged unsupported added content in $5$ units ($10\%$) that the judge's hallucination label did not flag. Three concern figures that cannot be checked against the reference, the weaker reading; the remaining two are substantive: the candidate asserts a weight-loss trend for a patient whose record contains no weight history, and one of the two was flagged independently by two raters. GPT-4o-mini scored those units $4$ on factuality and $5$ on safety. We therefore make no claim about the absence of hallucination, and the judge's hallucination label should not be used as a safety indicator: a reference-only comparison cannot see a claim the patient record silently fails to license.

\section{Conclusion}

IndoNutriQA fills a gap in Indonesian-language clinical QA with $911$ pairs written by three nutritionists and grounded in real clinical-nutrition reports written by graduate-level dietetic interns. The strongest results form a statistically tied top tier of $5$-shot LLMs (Llama-3.1-8B-Instruct F1 $= 55.55$, MedGemma-4B-IT $54.28$), above the best fine-tuned IndoBERT baseline by $+19.21$ F1, supporting IndoNutriQA as an evaluation resource. These conclusions are limited to the eight question types, the clinical reports, and the configurations tested. An expert-rater audit found $6$--$10\%$ of references contestable, safety scores unvalidated, and unsupported content the judge missed, so we claim no hallucination-free output. Future work will add cases from other facilities, extend the expert audit, and test stronger Indonesian generators for RAG.

\section*{Limitations}

This work has several limitations that should be considered when interpreting the results and using the IndoNutriQA dataset.

First, the size of the dataset is modest at $911$ question-answer pairs from $120$ source reports. This reflects the realistic constraints of collecting and annotating clinical data with clinical nutrition experts, and we therefore frame IndoNutriQA primarily as an evaluation benchmark rather than a large-scale training corpus (Sections~\ref{subsec:llm_evaluation}--\ref{subsec:statistical_robustness}). To address the statistical-power concern that follows from this scale, we report $5$-fold cross-validation stratified by source report (with \texttt{GroupKFold} forbidding any cross-report contamination), a learning-curve analysis over training-set fractions, and bootstrap $95\%$ confidence intervals on every headline metric, in addition to zero-shot and five-shot LLM evaluations that require no in-domain training data. Nevertheless, the modest scale limits the benchmark's ability to expose tail behaviours of data-hungry generative and retrieval-augmented systems; reported numbers should be interpreted as benchmark references rather than upper-bound performance estimates achievable with substantially more data.

A related concern is the scope of the question set. IndoNutriQA is built from eight fixed questions rather than free-form ones, which bounds what the benchmark can measure: it exercises the eight NCP/PES reasoning steps that structure Indonesian clinical-nutrition practice, and nothing outside them. It therefore does not test open-ended clinical dialogue, patient-initiated questions, or reasoning about sections of the report the eight questions do not touch (for instance pharmacological interactions or longitudinal monitoring, both present in the source documents but unqueried). We regard the fixed question set as a deliberate trade of coverage for control, since it makes answers comparable across $120$ patients and bounds inter-annotator variation, but it is a real ceiling on scope, and extending the question inventory is the most direct way to raise it.

A further concern is the number of source documents rather than the number of pairs: $911$ pairs drawn from $120$ reports average roughly $7.5$ pairs per report, so case diversity is bounded by the report count regardless of how many pairs are extracted. Three considerations bear on this. First, the eight questions are standardised and applied uniformly to every report, so the pairs within a report are not near-duplicates of one another but probe eight distinct sections of the NCP; the effective diversity of the benchmark is therefore closer to $120$ cases $\times$ $8$ reasoning types than to $911$ interchangeable items. Second, because the question side is fixed, annotator variation is confined to answer text, which is what makes a corpus of this size usable as a controlled evaluation instrument rather than merely a small training set. Third, the $120$ reports are not drawn from a single site or a single condition: they span multiple healthcare facilities and ten disease categories with no category exceeding $30\%$ of reports (Table~\ref{tab:disease_distribution}). None of this removes the ceiling that $120$ documents impose on rare-case coverage, which we return to below; it does mean the constraint is on case breadth rather than on redundancy among the pairs.

Second, the dataset exhibits domain and population skew. A substantial portion of the cases is derived from pediatric and respiratory-related clinical contexts, reflecting real-world hospital data availability, although no single disease category exceeds $30\%$ of source reports (Table~\ref{tab:disease_distribution}). We explicitly stratify performance by disease category in Section~\ref{subsec:stratified_analysis} (Table~\ref{tab:disease_breakdown}) and find that the dataset-level respiratory prominence does not translate to biased generalisation: Llama-3.1-8B (5-shot) ranges $50.08$--$64.17$ F1 across the seven categories represented in the test split, with respiratory ($57.55$) sitting within the spread. Two caveats remain. First, the held-out test split is not stratified by disease category: it over-represents neurologic ($20.0\%$ vs.\ $10.0\%$ of the corpus) and hematologic/oncologic ($15.0\%$ vs.\ $8.3\%$) while under-representing infectious ($5.0\%$ vs.\ $17.5\%$), and contains no cardiac or renal/urologic reports at all, so generalisation to those conditions remains untested on the held-out split. Second, the $5$-fold cross-validation of Section~\ref{subsec:statistical_robustness} uses \texttt{GroupKFold} over all $120$ reports and therefore does exercise every category, which is the stronger evidence for category-level robustness.

This bears directly on how far the reported results should be expected to generalise. Across institutions: the reports are authored at multiple Indonesian healthcare facilities spanning public regional hospitals, private medical centres, and a military hospital, rather than at a single site. This matters because clinical-nutrition documentation is shaped by institutional templates, and a benchmark drawn from one hospital would reward models that learn that hospital's report layout. Two design choices limit that risk: the eight standardised questions derive from the NCP and PES frameworks, which are national practice standards rather than site conventions, and the cross-validation groups by source report so that no institutional template appears in both training and evaluation folds, with fold-level standard deviations remaining small ($\le 2.66$ F1). What remains untested is transfer to facilities outside the contributing set, particularly primary-care puskesmas settings, where documentation is briefer and less standardised than the inpatient reports collected here. Across rare conditions: renal/urologic and ENT cases each account for under $1\%$ of source reports, and for conditions represented by one or two reports no partition can support a reliable per-category estimate. We would expect degradation on conditions whose nutritional management diverges sharply from the modal pediatric case, for instance renal diets with protein and electrolyte restriction, which invert the high-protein prescriptions dominating this corpus. Extending coverage of these conditions is the highest-value direction for a second data-collection round, and we flag it explicitly so that users of the benchmark do not over-read aggregate scores as evidence of broad clinical competence.

Third, single-annotator pairs do not guarantee a unique gold answer per question, particularly for the recommendation-style questions. We mitigate this on two axes: the NCP/PES guideline structurally constrains acceptable answers (Appendix~\ref{subsec:question_types}), and our evaluation protocol uses paraphrase-tolerant semantic and rubric metrics (Section~\ref{subsec:semantic_evaluation}) alongside lexical scores. We also report inter-rater agreement directly: three expert raters independently re-scored $50$ stratified test-split units (Appendix~\ref{appx:rd_judge}), reaching quadratic-weighted $\kappa = 0.780$ on factuality and $0.752$ on completeness but only $0.160$ on clinical safety. The same exercise flagged $3$ of $50$ reference answers ($6\%$) as incorrect or contestable by explicit flag, rising to $5$ ($10\%$) if free-text criticism of the reference is counted (Section~\ref{sec:annotation}); we take this as a direct estimate of gold-answer error in the test split, and those items are released with the benchmark so they can be excluded or re-adjudicated. Because two of the three raters also wrote references, this estimate is not fully independent. Extending this re-annotation beyond $50$ units, and to the training split, remains future work.

Fourth, all data are sourced from Indonesian healthcare facilities and written in Indonesian. While this is a deliberate design choice to support low-resource clinical NLP, it restricts direct applicability to other languages and healthcare systems without additional adaptation. The benchmark does provide partial cross-lingual signals, since two large multilingual encoders (XLM-R-large, mDeBERTa-v3-base) are fine-tuned on the Indonesian training data, and two English medical LLMs (BioMistral-7B, MedGemma-4B-IT) are evaluated zero- and five-shot on Indonesian clinical text, but a parallel English-translated test split for fully aligned cross-lingual experiments is left for future work.

Fifth, although the evaluation protocol goes beyond lexical-overlap metrics (Section~\ref{subsec:semantic_evaluation} adds BERTScore, sentence-embedding cosine, and an LLM-as-judge clinical rubric, with a second-judge cross-validation using Claude Haiku 4.5 yielding pooled $\kappa = 0.652$), per-prediction human evaluation by expert raters was conducted on a $50$-unit stratified sample rather than at full scale (Appendix~\ref{appx:rd_judge}). That calibration validates the rubric unevenly. On factuality and completeness, judge-vs-rater agreement ($\kappa = 0.705$ and $0.623$ for GPT-4o-mini) approaches the inter-rater ceiling ($0.780$ and $0.752$), so on those dimensions the judge is about as close to a rater as raters are to each other. On clinical safety it does not: agreement is $\kappa = 0.133$ (GPT-4o-mini) and $0.480$ (Claude Haiku 4.5), and both judges score safety above the rater consensus, by $+1.14$ and $+0.58$ rubric points respectively. The raters also disagree among themselves on safety (Krippendorff $\alpha = 0.250$), so this is not solely a judge failure: safety as operationalised here lacks a stable human target. We therefore report the safety column as unvalidated, and no safety score in this paper should be read as a clinical guarantee. At $50$ units the per-stratum intervals remain wide, and agreement on a stratified sample need not transfer to the full test split. The human error analysis derived from the same exercise (Section~\ref{subsec:human_error}) carries its own limits: the raters' notes were optional and unevenly distributed across the three raters, so its category counts are lower bounds on prevalence rather than error rates, and the coding of those notes into categories is ours rather than the raters' (Appendix~\ref{appx:human_error}). It is nonetheless the finding that most changes how our own results should be read, since it shows the judge's hallucination label to be a detection ceiling rather than a measure of unsupported clinical content. The primary judge (GPT-4o-mini) also provides two evaluated systems; a difference-in-differences test against a control judge from a different provider finds no detectable self-preference (Appendix~\ref{appx:judge_bias}), but with four same-provider systems that test is underpowered, and it cannot rule out a stylistic prior shared by both judges.

Finally, the retrieval-augmented generation (RAG) baseline exhibits low lexical scores. The RAGAS diagnostic in Section~\ref{subsec:ragas_evaluation} localises this to the \texttt{mt5-base} generator (faithfulness $12.86$ on the $105$ of $137$ examples where it is defined, answer relevancy $0.24$) while the retriever is the stronger component (context precision $73.43$). Because we evaluate a single retriever--generator--chunking configuration, these numbers should be read as a localisation of failure within that configuration and not as evidence about the effectiveness of RAG for Indonesian clinical QA in general. A stronger Indonesian generator with grounding-aware training is the most direct next step, and retriever-side changes (recall is only $35.98$) were not tested; both are left for future work.

Despite these limitations, IndoNutriQA provides a clinically grounded and ethically curated resource that establishes a foundation for future research in Indonesian clinical question answering and nutrition-focused decision support.

\section*{Ethics Statement}

This work involves the construction and release of a clinical question answering dataset derived from real-world clinical nutrition reports. Ethical considerations related to data privacy, consent, potential harm, and responsible use were carefully addressed throughout the dataset development and release process.

\noindent\textbf{Patient-level anonymisation.}~All source documents were fully anonymised prior to annotation and release. Personally identifiable information (PII), including patient names, medical record numbers, registration numbers, addresses, and dates of birth, was removed using rule-based procedures tailored to Indonesian clinical documents and replaced with deterministic placeholders (\texttt{<NAMA>}, \texttt{<NO\_RM>}, \texttt{<NO\_REG>}, \texttt{<TGL>}, \texttt{<LOKASI>}). As a result, the released dataset does not contain information that can be used to identify individual patients. No attempts were made to re-identify individuals, and the dataset is intended solely for research and educational purposes.

\noindent\textbf{Institution-level anonymisation.}~A second anonymisation pass strips identifying institution mentions that the per-chunk NER list would not catch but that nevertheless identify the host site: hospital names (\texttt{<RS>}, e.g., regional public hospitals, private medical centers, military naval hospitals, and other named or acronym hospital references) and publisher names attached to academic citations (\texttt{<PENERBIT>}, e.g., medical textbook publishers, university presses). The two passes together leave the released dataset stripped of both individual-patient and host-institution identifiers.

\noindent\textbf{Annotation and consent.}~The dataset was constructed from retrospective clinical-nutrition reports authored by graduate-level dietetic interns as part of supervised clinical nutrition rotations. No direct patient contact or intervention was involved in this study. Each (report, question, answer) triple has exactly one author, one of three nutritionists who wrote the answers to reports that graduate-level dietetic interns had written under lecturer supervision (Appendix~\ref{appx:annotation}). The study does not include vulnerable-population targeting beyond what is inherent in clinical nutrition practice (e.g., pediatric cases), and no demographic attributes were collected or inferred beyond those explicitly documented in anonymised clinical contexts.

\noindent\textbf{Intended use.}~IndoNutriQA is released with the explicit intention of supporting research in clinical natural language processing and low-resource language question answering. The dataset is not intended for direct clinical deployment or automated medical decision making. Any predictions or recommendations generated by models trained or evaluated on this dataset should be interpreted as decision-support signals and must not replace professional medical judgement.

\noindent\textbf{Risks and mitigations.}~Potential risks include the misuse of trained models for unsupported clinical inference or deployment without appropriate validation. To mitigate this: (i) the dataset documentation states usage limitations explicitly; (ii) benchmark results are reported as baseline references with bootstrap confidence intervals rather than performance guarantees (Section~\ref{subsec:statistical_robustness}); (iii) we make no claim that the evaluated systems are free of hallucination: the LLM judge's hallucination label is unvalidated, and expert raters auditing the same kind of output independently identified unsupported clinical additions in $10\%$ of units, two of them asserting findings the patient record does not license (Section~\ref{subsec:human_error}); and (iv) the LLM-as-judge clinical rubric is cross-validated across two judge families (GPT-4o-mini and Claude Haiku 4.5, pooled Cohen's $\kappa = 0.652$; Section~\ref{subsec:semantic_evaluation}). Researchers using IndoNutriQA are nevertheless encouraged to conduct additional validation, including expert review by qualified clinicians, before applying derived models in real-world clinical settings.

Overall, this work adheres to established ethical guidelines for clinical NLP research, prioritising patient privacy, institutional anonymity, transparency, and responsible use while contributing a valuable resource for advancing medical and nutritional decision support in low-resource language contexts.

\paragraph{Data and Code Availability.} The anonymized IndoNutriQA dataset and the full evaluation pipeline (including the prediction caches needed to reproduce every table and figure in this paper without re-querying LLM APIs) are publicly released at \url{https://github.com/NLP-UB/IndoNutriQA}. The dataset is released under CC BY-NC 4.0, and the code is released under the MIT license.

\section*{Acknowledgements}
This research was funded by Universitas Brawijaya. We thank the lecturers who supervised the reports and the graduate-level dietetic interns whose clinical-nutrition reports made this dataset possible. We also thank Al Fila Miftahul Jannah, Firda Aqilatul 'Izzah and Reynata Olivia Kristabel, the nutritionists who wrote the question answers, two of whom also took part in the validation.

\bibliography{references}

@inproceedings{rajpurkar2016squad,
  title     = {{SQ}u{AD}: 100,000+ Questions for Machine Comprehension of Text},
  author    = {Rajpurkar, Pranav  and
               Zhang, Jian  and
               Lopyrev, Konstantin  and
               Liang, Percy},
  editor    = {Su, Jian  and
               Duh, Kevin  and
               Carreras, Xavier},
  booktitle = {Proceedings of the 2016 Conference on Empirical Methods in Natural Language Processing},
  month     = nov,
  year      = {2016},
  address   = {Austin, Texas},
  publisher = {Association for Computational Linguistics},
  url       = {https://aclanthology.org/D16-1264/},
  doi       = {10.18653/v1/D16-1264},
  pages     = {2383--2392}
}

@article{kwiatkowski2019natural,
  title     = {Natural Questions: A Benchmark for Question Answering Research},
  author    = {Kwiatkowski, Tom  and
               Palomaki, Jennimaria  and
               Redfield, Olivia  and
               Collins, Michael  and
               Parikh, Ankur  and
               Alberti, Chris  and
               Epstein, Danielle  and
               Polosukhin, Illia  and
               Devlin, Jacob  and
               Lee, Kenton  and
               Toutanova, Kristina  and
               Jones, Llion  and
               Kelcey, Matthew  and
               Chang, Ming-Wei  and
               Dai, Andrew M.  and
               Uszkoreit, Jakob  and
               Le, Quoc  and
               Petrov, Slav},
  editor    = {Lee, Lillian  and
               Johnson, Mark  and
               Roark, Brian  and
               Nenkova, Ani},
  journal   = {Transactions of the Association for Computational Linguistics},
  volume    = {7},
  year      = {2019},
  address   = {Cambridge, MA},
  publisher = {MIT Press},
  url       = {https://aclanthology.org/Q19-1026/},
  doi       = {10.1162/tacl_a_00276},
  pages     = {453--466}
}

@article{clark2020tydiqa,
  title     = {{T}y{D}i {QA}: A Benchmark for Information-Seeking Question Answering in Typologically Diverse Languages},
  author    = {Clark, Jonathan H.  and
               Choi, Eunsol  and
               Collins, Michael  and
               Garrette, Dan  and
               Kwiatkowski, Tom  and
               Nikolaev, Vitaly  and
               Palomaki, Jennimaria},
  editor    = {Johnson, Mark  and
               Roark, Brian  and
               Nenkova, Ani},
  journal   = {Transactions of the Association for Computational Linguistics},
  volume    = {8},
  year      = {2020},
  address   = {Cambridge, MA},
  publisher = {MIT Press},
  url       = {https://aclanthology.org/2020.tacl-1.30/},
  doi       = {10.1162/tacl_a_00317},
  pages     = {454--470}
}

@article{tsatsaronis2015overview,
  author  = {George Tsatsaronis and Georgios Balikas and Prodromos Malakasiotis and Ioannis Partalas and Matthias Zschunke and Michael R Alvers and Dirk Weissenborn and Anastasia Krithara and Sergios Petridis and Dimitris Polychronopoulos and Yannis Almirantis and John Pavlopoulos and Nicolas Baskiotis and Patrick Gallinari and Thierry Artiéres and Axel-Cyrille Ngonga Ngomo and Norman Heino and Eric Gaussier and Liliana Barrio-Alvers and Michael Schroeder and Ion Androutsopoulos and Georgios Paliouras},
  doi     = {10.1186/s12859-015-0564-6},
  issn    = {1471-2105},
  issue   = {1},
  journal = {BMC Bioinformatics},
  month   = {12},
  pages   = {138},
  title   = {An overview of the BIOASQ large-scale biomedical semantic indexing and question answering competition},
  volume  = {16},
  year    = {2015}
}

@inproceedings{wilie2020indonlu,
  title     = {{I}ndo{NLU}: Benchmark and Resources for Evaluating {I}ndonesian Natural Language Understanding},
  author    = {Wilie, Bryan  and
               Vincentio, Karissa  and
               Winata, Genta Indra  and
               Cahyawijaya, Samuel  and
               Li, Xiaohong  and
               Lim, Zhi Yuan  and
               Soleman, Sidik  and
               Mahendra, Rahmad  and
               Fung, Pascale  and
               Bahar, Syafri  and
               Purwarianti, Ayu},
  editor    = {Wong, Kam-Fai  and
               Knight, Kevin  and
               Wu, Hua},
  booktitle = {Proceedings of the 1st Conference of the Asia-Pacific Chapter of the Association for Computational Linguistics and the 10th International Joint Conference on Natural Language Processing},
  month     = dec,
  year      = {2020},
  address   = {Suzhou, China},
  publisher = {Association for Computational Linguistics},
  url       = {https://aclanthology.org/2020.aacl-main.85/},
  doi       = {10.18653/v1/2020.aacl-main.85},
  pages     = {843--857}
}

@inproceedings{cahyawijaya2021indonlg,
  title     = {{I}ndo{NLG}: Benchmark and Resources for Evaluating {I}ndonesian Natural Language Generation},
  author    = {Cahyawijaya, Samuel  and
               Winata, Genta Indra  and
               Wilie, Bryan  and
               Vincentio, Karissa  and
               Li, Xiaohong  and
               Kuncoro, Adhiguna  and
               Ruder, Sebastian  and
               Lim, Zhi Yuan  and
               Bahar, Syafri  and
               Khodra, Masayu  and
               Purwarianti, Ayu  and
               Fung, Pascale},
  editor    = {Moens, Marie-Francine  and
               Huang, Xuanjing  and
               Specia, Lucia  and
               Yih, Scott Wen-tau},
  booktitle = {Proceedings of the 2021 Conference on Empirical Methods in Natural Language Processing},
  month     = nov,
  year      = {2021},
  address   = {Online and Punta Cana, Dominican Republic},
  publisher = {Association for Computational Linguistics},
  url       = {https://aclanthology.org/2021.emnlp-main.699/},
  doi       = {10.18653/v1/2021.emnlp-main.699},
  pages     = {8875--8898}
}

@inproceedings{koto2020indolem,
  title     = {{I}ndo{LEM} and {I}ndo{BERT}: A Benchmark Dataset and Pre-trained Language Model for {I}ndonesian {NLP}},
  author    = {Koto, Fajri  and
               Rahimi, Afshin  and
               Lau, Jey Han  and
               Baldwin, Timothy},
  editor    = {Scott, Donia  and
               Bel, Nuria  and
               Zong, Chengqing},
  booktitle = {Proceedings of the 28th International Conference on Computational Linguistics},
  month     = dec,
  year      = {2020},
  address   = {Barcelona, Spain (Online)},
  publisher = {International Committee on Computational Linguistics},
  url       = {https://aclanthology.org/2020.coling-main.66/},
  doi       = {10.18653/v1/2020.coling-main.66},
  pages     = {757--770}
}

@article{jin2021disease,
  author   = {Di Jin and Eileen Pan and Nassim Oufattole and Wei-Hung Weng and Hanyi Fang and Peter Szolovits},
  doi      = {10.3390/app11146421},
  issn     = {2076-3417},
  issue    = {14},
  journal  = {Applied Sciences},
  month    = {7},
  pages    = {6421},
  title    = {What Disease Does This Patient Have? A Large-Scale Open Domain Question Answering Dataset from Medical Exams},
  volume   = {11},
  year     = {2021}
}

@inproceedings{pal2022medmcqa,
  title     = {MedMCQA: A Large-scale Multi-Subject Multi-Choice Dataset for Medical domain Question Answering},
  author    = {Pal, Ankit and Umapathi, Logesh Kumar and Sankarasubbu, Malaikannan},
  booktitle = {Proceedings of the Conference on Health, Inference, and Learning},
  pages     = {248--260},
  year      = {2022},
  editor    = {Flores, Gerardo and Chen, George H and Pollard, Tom and Ho, Joyce C and Naumann, Tristan},
  volume    = {174},
  series    = {Proceedings of Machine Learning Research},
  month     = {07--08 Apr},
  publisher = {PMLR},
  url       = {https://proceedings.mlr.press/v174/pal22a.html}
}

@inproceedings{jin2019pubmedqa,
  title     = {{P}ub{M}ed{QA}: A Dataset for Biomedical Research Question Answering},
  author    = {Jin, Qiao  and
               Dhingra, Bhuwan  and
               Liu, Zhengping  and
               Cohen, William  and
               Lu, Xinghua},
  editor    = {Inui, Kentaro  and
               Jiang, Jing  and
               Ng, Vincent  and
               Wan, Xiaojun},
  booktitle = {Proceedings of the 2019 Conference on Empirical Methods in Natural Language Processing and the 9th International Joint Conference on Natural Language Processing (EMNLP-IJCNLP)},
  month     = nov,
  year      = {2019},
  address   = {Hong Kong, China},
  publisher = {Association for Computational Linguistics},
  url       = {https://aclanthology.org/D19-1259/},
  doi       = {10.18653/v1/D19-1259},
  pages     = {2567--2577}
}

@inproceedings{pampari2018emrqa,
  title     = {emr{QA}: A Large Corpus for Question Answering on Electronic Medical Records},
  author    = {Pampari, Anusri  and
               Raghavan, Preethi  and
               Liang, Jennifer  and
               Peng, Jian},
  editor    = {Riloff, Ellen  and
               Chiang, David  and
               Hockenmaier, Julia  and
               Tsujii, Jun{'}ichi},
  booktitle = {Proceedings of the 2018 Conference on Empirical Methods in Natural Language Processing},
  month     = oct # {-} # nov,
  year      = {2018},
  address   = {Brussels, Belgium},
  publisher = {Association for Computational Linguistics},
  url       = {https://aclanthology.org/D18-1258/},
  doi       = {10.18653/v1/D18-1258},
  pages     = {2357--2368}
}

@article{abacha2019medquad,
  author  = {Asma Ben Abacha and Dina Demner-Fushman},
  doi     = {10.1186/s12859-019-3119-4},
  issn    = {1471-2105},
  issue   = {1},
  journal = {BMC Bioinformatics},
  month   = {12},
  pages   = {511},
  title   = {A question-entailment approach to question answering},
  volume  = {20},
  year    = {2019}
}

@inproceedings{kim2024medexqa,
  title     = {{M}ed{E}x{QA}: Medical Question Answering Benchmark with Multiple Explanations},
  author    = {Kim, Yunsoo  and
               Wu, Jinge  and
               Abdulle, Yusuf  and
               Wu, Honghan},
  editor    = {Demner-Fushman, Dina  and
               Ananiadou, Sophia  and
               Miwa, Makoto  and
               Roberts, Kirk  and
               Tsujii, Junichi},
  booktitle = {Proceedings of the 23rd Workshop on Biomedical Natural Language Processing},
  month     = aug,
  year      = {2024},
  address   = {Bangkok, Thailand},
  publisher = {Association for Computational Linguistics},
  url       = {https://aclanthology.org/2024.bionlp-1.14/},
  doi       = {10.18653/v1/2024.bionlp-1.14},
  pages     = {167--181}
}

@article{demnerfushman2007clinical,
  author   = {Demner-Fushman, Dina and Lin, Jimmy},
  title    = {Answering Clinical Questions with Knowledge-Based and Statistical Techniques},
  journal  = {Computational Linguistics},
  volume   = {33},
  number   = {1},
  pages    = {63-103},
  year     = {2007},
  month    = {03},
  issn     = {0891-2017},
  doi      = {10.1162/coli.2007.33.1.63},
  url      = {https://doi.org/10.1162/coli.2007.33.1.63},
  eprint   = {https://direct.mit.edu/coli/article-pdf/33/1/63/1798375/coli.2007.33.1.63.pdf}
}

@article{mutabazi2021review,
  author         = {Mutabazi, Emmanuel and Ni, Jianjun and Tang, Guangyi and Cao, Weidong},
  title          = {A Review on Medical Textual Question Answering Systems Based on Deep Learning Approaches},
  journal        = {Applied Sciences},
  volume         = {11},
  year           = {2021},
  number         = {12},
  article-number = {5456},
  url            = {https://www.mdpi.com/2076-3417/11/12/5456},
  issn           = {2076-3417},
  doi            = {10.3390/app11125456}
}

@article{lee2020biobert,
  author   = {Lee, Jinhyuk and Yoon, Wonjin and Kim, Sungdong and Kim, Donghyeon and Kim, Sunkyu and So, Chan Ho and Kang, Jaewoo},
  title    = {{B}io{BERT}: a pre-trained biomedical language representation model for biomedical text mining},
  journal  = {Bioinformatics},
  volume   = {36},
  number   = {4},
  pages    = {1234-1240},
  year     = {2020},
  issn     = {1367-4803},
  doi      = {10.1093/bioinformatics/btz682},
  url      = {https://doi.org/10.1093/bioinformatics/btz682},
  eprint   = {https://academic.oup.com/bioinformatics/article-pdf/36/4/1234/48983216/bioinformatics_36_4_1234.pdf}
}

@inproceedings{alsentzer2019publicly,
  author    = {Emily Alsentzer and John Murphy and William Boag and Wei-Hung Weng and Di Jindi and Tristan Naumann and Matthew McDermott},
  city      = {Stroudsburg, PA, USA},
  doi       = {10.18653/v1/W19-1909},
  booktitle = {Proceedings of the 2nd Clinical Natural Language Processing Workshop},
  pages     = {72-78},
  publisher = {Association for Computational Linguistics},
  title     = {Publicly Available Clinical {BERT} Embeddings},
  year      = {2019}
}

@techreport{kemenkes2018riskesdas,
  title       = {Laporan Nasional Riset Kesehatan Dasar (RISKESDAS) 2018 (National Report on Basic Health Research)},
  author      = {{Kementerian Kesehatan Republik Indonesia}},
  institution = {Badan Penelitian dan Pengembangan Kesehatan},
  year        = {2019},
  address     = {Jakarta, Indonesia},
  url       = {https://www.badankebijakan.kemkes.go.id/repositori/id/eprint/3514/}
}

@article{Oktaria2017,
  author  = {V. Oktaria and K. J. Lee and J. E. Bines and E. Watts and C. D. Satria and J. Atthobari and H. Nirwati and C. D. Kirkwood and Y. Soenarto and M. H. Danchin},
  doi     = {10.1186/s12887-017-0966-x},
  issn    = {1471-2431},
  issue   = {1},
  journal = {BMC Pediatrics},
  month   = {12},
  pages   = {211},
  title   = {Nutritional status, exclusive breastfeeding and management of acute respiratory illness and diarrhea in the first 6 months of life in infants from two regions of Indonesia},
  volume  = {17},
  year    = {2017}
}

@article{Vihikan2024,
  author   = {Wayan Oger Vihikan and I Nyoman Prayana Trisna},
  doi      = {10.51519/journalisi.v6i3.838},
  issn     = {2656-4882},
  issue    = {3},
  journal  = {Journal of Information Systems and Informatics},
  month    = {9},
  pages    = {1931-1944},
  title    = {Indonesian Health Question Multi-Class Classification Based on Deep Learning},
  volume   = {6},
  year     = {2024}
}

@inproceedings{Aji2021,
  author    = {Alham Fikri Aji and Haryo Akbarianto Wibowo and Made Nindyatama Nityasya and Radityo Eko Prasojo and Tirana Noor Fatyanosa},
  booktitle = {The Social Media Mining for Health Applications (SMM4H) workshop},
  title     = {{BERT} Goes Brrr: A Venture Towards the Lesser Error in Classifying Medical Self-Reporters on Twitter},
  year      = {2021},
  url       = {https://aclanthology.org/2021.smm4h-1.9}
}

@inproceedings{conneau2020xlmr,
  author    = {Alexis Conneau and Kartikay Khandelwal and Naman Goyal and Vishrav Chaudhary and Guillaume Wenzek and Francisco Guzm{\'{a}}n and Edouard Grave and Myle Ott and Luke Zettlemoyer and Veselin Stoyanov},
  title     = {Unsupervised Cross-Lingual Representation Learning at Scale},
  booktitle = {Proceedings of the 58th Annual Meeting of the Association for Computational Linguistics},
  pages     = {8440--8451},
  year      = {2020},
  url       = {https://aclanthology.org/2020.acl-main.747/}
}

@inproceedings{es2024ragas,
  author    = {Shahul Es and Jithin James and Luis Espinosa Anke and Steven Schockaert},
  title     = {{RAGAS}: Automated Evaluation of Retrieval Augmented Generation},
  booktitle = {Proceedings of the 18th Conference of the European Chapter of the Association for Computational Linguistics: System Demonstrations},
  year      = {2024},
  url       = {https://aclanthology.org/2024.eacl-demo.16/}
}

@inproceedings{zhang2020bertscore,
  author    = {Tianyi Zhang and Varsha Kishore and Felix Wu and Kilian Q. Weinberger and Yoav Artzi},
  title     = {{BERTScore}: Evaluating Text Generation with {BERT}},
  booktitle = {International Conference on Learning Representations},
  year      = {2020},
  url       = {https://arxiv.org/abs/1904.09675}
}

\appendix

\section{Annotation and Context-Pairing Details}
\label{appx:annotation}

\subsection{Question Types and Complexity}
\label{subsec:question_types}

The questions in IndoNutriQA are designed to reflect real-world clinical nutrition workflows and span multiple levels of reasoning complexity. Overall, the dataset includes four major categories of questions. Factual questions target explicit clinical information such as anthropometric measurements, laboratory values, or diagnostic outcomes. Inferential questions require clinical reasoning by integrating multiple pieces of contextual information, for example combining anthropometric findings with medical diagnoses. Comparative questions involve assessing differences across parameters or time points, such as changes in nutritional status or laboratory indicators. Finally, recommendation questions focus on clinical decision support, asking for appropriate dietary goals, principles, or guidelines based on the patient’s condition.

To ensure consistency and comprehensive coverage across cases, each patient record is associated with a standardized set of NCP/PES questions that operationalize these categories. The complete list of standardized questions applied uniformly to all cases is provided in Appendix \ref{sec:appendix}. These questions collectively cover factual assessment, clinical reasoning, and recommendation-oriented decision making, ensuring that models trained on IndoNutriQA are evaluated across a broad spectrum of clinically relevant clinical question answering scenarios. Table~\ref{tab:statistics} gives the main dataset statistics, including the average question, answer, and context lengths.

\begin{table}[!tbp]

\centering
\resizebox{0.8\columnwidth}{!}{

\begin{tabular}{@{}ll@{}}
\toprule
\textbf{Metric} & \textbf{Value} \\ \midrule
Total QA Pairs & 911 \\
Total Source Reports & 120 \\
Average Question Length & 8.32 tokens \\
Average Answer Length & 70.66 tokens \\
Average Context Length & 131.28 tokens \\
Unique Medical Diagnoses & 128 \\
Average Chunks per Report & 2.60 \\ \bottomrule
\end{tabular}}
\caption{Dataset Statistics Overview}
\label{tab:statistics}
\end{table}

\subsection*{Annotator Pool}
The $120$ source reports were written by graduate-level dietetic interns at the Faculty of Health Sciences, Universitas Brawijaya, as inpatient or sub-inpatient clinical rotation deliverables of a dietetic internship. Each report was supervised by a lecturer of the faculty. The reports are the source material, not the annotation. The question answers were written by three nutritionists, each assigned a different set of reports. Each nutritionist wrote the eight answers for every report in their set.

\subsection*{Per-pair Annotation Count and Validation}
Each (report, question) pair is answered by exactly one nutritionist. We do not collect multiple independent annotations of the same pair, because adding parallel annotators would duplicate the entire chart-review workflow per report and was outside the scope of this dataset release. A stratified sample of $50$ units of model output was later re-scored by one Registered Dietitian, who took no part in writing the answers, and two of the three nutritionists (Appendix~\ref{appx:rd_judge}). That study provides the only validation of the reference answers; we do not claim that all pairs were reviewed.

\subsection*{Annotation Guideline}
The nutritionists followed the clinical-nutrition documentation guideline used in the clinical nutrition rotations, which is grounded in two standardized clinical-nutrition reference frameworks: the Nutrition Care Process (NCP) for clinical-nutrition documentation, and the PES (Problem--Etiology--Sign/Symptom) format for nutrition diagnosis statements. The eight standardized questions correspond directly to NCP/PES sections (screening, anthropometric assessment, biochemical assessment, nutrition diagnosis, dietary goals, dietary principles, dietary requirements, dietary guidelines), so the answer space for each question is bounded by the NCP/PES rubric rather than by free-form clinical opinion.

\subsection*{Subjectivity and Inter-Annotator Agreement}
\label{subsec:annotation_subjectivity}
We do not claim that a single nutritionist's annotation produces a unique gold answer for every case: clinically valid alternatives exist, particularly for the recommendation-style questions. Two design choices mitigate this. First, the NCP/PES guideline structurally constrains acceptable answers: a PES diagnosis (Question~4) must contain the three required components in the prescribed order, and a dietary-goals statement (Question~5) is bounded by the patient's clinical condition and standard-of-care guidelines. Second, our experimental protocol uses metrics that are robust to clinically valid lexical variation: BERTScore, sentence-embedding cosine similarity, and the LLM-as-judge rubric of Section~\ref{subsec:semantic_evaluation} explicitly do not penalize paraphrase, and the LLM judge scores factuality and clinical safety separately. Three caveats remain, the first now partly resolved by measurement rather than argument. (i) A conventional inter-annotator $\kappa$ over the references is undefined here, because each report was answered by a single nutritionist; in its place we report the three-rater re-scoring study and the gold-answer audit of Section~\ref{sec:annotation} and Appendix~\ref{appx:rd_judge}, which give an expert-agreement ceiling (quadratic $\kappa = 0.780$ factuality, $0.752$ completeness, $0.160$ safety) and a direct reference-error estimate ($6\%$ by explicit flag, $10\%$ including free-text criticism). Extending that audit beyond $50$ units, and to the training split, remains future work. (ii) Two of the three raters also wrote references, so the audit is not fully independent: only the RD took no part in writing the answers. (iii) The three-nutritionist pipeline is one specific protocol; reproducing this dataset under a parallel multi-annotator protocol would establish a strict upper bound on what single-annotator authoring can recover, and because each nutritionist wrote answers for different reports, author effects cannot be separated from report effects.

\subsection*{Question-Context Pairing: Six-Stage Matching}
Unlike SQuAD-style datasets in which an annotator writes both the question and a span within a fixed passage, IndoNutriQA's context field is algorithmically located from the source clinical report after the answer has been authored. For each (report, question, answer) triple we assign the \texttt{context} field to the chunk in which the authored answer is best located, using a six-stage matching strategy applied in order: (1) exact substring match; (2) case-insensitive match; (3) whitespace-normalized match (collapsing multi-space and newline variants); (4) case-insensitive whitespace-normalized match; (5) punctuation-stripped match; (6) sliding-window fragment matching with progressively shorter prefixes of the answer ($\{150, 100, 80, 60, 40, 30, 20\}$ characters). The character-level start position of the answer within the chunk is recorded as \texttt{answer\_start}, supporting extractive-QA training under a SQuAD-compatible schema. Across the full $911$-pair release, the procedure locates a verified character-level position in $823$ pairs ($90.3\%$); the remaining $88$ pairs ($9.7\%$) are released with a non-negative \texttt{chunk\_index} pointing to the most similar chunk but without a verified character-level \texttt{answer\_start}, and are flagged in the released metadata so downstream users requiring strictly extractive-style supervision can filter on this column.

\section{Statistical Robustness Details}
\label{appx:statistical}

\subsection*{Statistical Robustness: full discussion}

To address the statistical-power concern from $n = 137$, we run three robustness procedures: $5$-fold cross-validation stratified by source report, a learning-curve analysis over training-set fractions $\{10, 25, 50, 100\}\%$, and bootstrap $95\%$ CIs over $1{,}000$ resamples on every reported metric. Cross-validated F1 for the best extractive baseline ($32.62 \pm 1.29$) is consistent with its held-out estimate ($36.34$, $95\%$ CI $[29.03, 43.59]$), with uniformly small fold-level standard deviations ($\le 2.66$ F1), so rankings are stable across partitions. The learning curve localizes bottlenecks: Rifky\_IndoBERT\_QA still rises at $100\%$ training data (data-bound), while mT5 plateaus near zero (architecture-bound). Bootstrap CIs show the top three $5$-shot LLMs (Llama-3.1-8B, MedGemma-4B-IT, GPT-4o) cluster within overlapping intervals, so the headline ranking is a top tier rather than a strict ordering, sitting well above the best fine-tuned baseline (non-overlapping CIs; Appendix~\ref{appx:statistical}).

\subsection*{Bootstrap $95\%$ Confidence Intervals}
Table~\ref{tab:bootstrap_ci} reports F1 with bootstrap $95\%$ confidence intervals over $1{,}000$ resamples of the test set ($n = 137$) for every model in Table~\ref{tab:model_performance}. CIs overlap among the top three 5-shot LLMs, supporting the tier-rather-than-ordering interpretation discussed in Section~\ref{subsec:statistical_robustness}.

\begin{table}[!tbp]
\centering

\resizebox{0.8\columnwidth}{!}{
\begin{tabular}{lc}
\hline
\textbf{Model} & \textbf{F1 \% [95\% CI]} \\
\hline
\multicolumn{2}{l}{\textit{Fine-tuned baselines}} \\
Rifky\_IndoBERT\_QA       & 36.34 [29.03, 43.59] \\
IndoLEM\_IndoBERT         & 23.78 [17.48, 29.75] \\
Cahya\_IndoBERT           & 16.56 [11.20, 21.96] \\
XLM-R-large               & 12.60 [ 8.05, 17.26] \\
mDeBERTa-v3-base          & 19.35 [13.34, 25.42] \\
mT5 (Generative)          &  7.84 [ 5.82,  9.92] \\
\hline
\multicolumn{2}{l}{\textit{Closed-source LLM}} \\
GPT-4o-mini (0-shot)      & 36.06 [32.22, 40.16] \\
GPT-4o-mini (5-shot)      & 43.07 [38.75, 47.46] \\
GPT-4o (0-shot)           & 37.49 [32.60, 42.22] \\
GPT-4o (5-shot)           & 50.33 [45.14, 55.58] \\
\hline
\multicolumn{2}{l}{\textit{Open-source multilingual LLM}} \\
Llama-3.1-8B (0-shot)     & 37.82 [33.76, 42.27] \\
Llama-3.1-8B (5-shot)     & \textbf{55.55} [49.96, 60.78] \\
Qwen2.5-7B (0-shot)       & 31.03 [27.21, 34.79] \\
Qwen2.5-7B (5-shot)       & 42.62 [37.72, 47.21] \\
\hline
\multicolumn{2}{l}{\textit{Indonesian-tuned LLM}} \\
Sahabat-AI-Llama3-8B (0-shot)  & 26.78 [23.22, 30.35] \\
Sahabat-AI-Llama3-8B (5-shot)  & 43.32 [37.44, 48.79] \\
Merak-7B-v4 (0-shot)      & 32.44 [28.40, 36.66] \\
Merak-7B-v4 (5-shot)      & 41.83 [37.00, 46.75] \\
\hline
\multicolumn{2}{l}{\textit{English-medical LLM}} \\
BioMistral-7B (0-shot)    & 29.83 [25.13, 34.79] \\
BioMistral-7B (5-shot)    & 40.08 [35.42, 45.15] \\
MedGemma-4B-IT (0-shot)   & 42.60 [38.14, 46.85] \\
MedGemma-4B-IT (5-shot)   & 54.28 [49.02, 59.42] \\
\hline
\end{tabular}}
\caption{F1 with bootstrap $95\%$ confidence intervals over $1{,}000$ resamples of the test set ($n = 137$). LLM rows distinguish 0-shot and 5-shot. Bold marks the best point estimate.}
\label{tab:bootstrap_ci}
\end{table}

\subsection*{$5$-Fold Cross-Validation (Stratified by Source Report)}
Each fine-tuned baseline is re-trained under \texttt{GroupKFold} cross-validation so that chunks from a single clinical document never appear in both training and evaluation folds. Table~\ref{tab:cv_results} reports per-fold mean and sample standard deviation. The cross-validated mean for the best extractive baseline ($32.62 \pm 1.29$ F1) is consistent with the held-out test estimate of $36.34$ (Table~\ref{tab:model_performance}).

\begin{table}[!tbp]
\centering

\resizebox{\columnwidth}{!}{
\begin{tabular}{lccc}
\hline
\textbf{Model} & \textbf{EM (\%)} & \textbf{F1 (\%)} & \textbf{ROUGE-L (\%)} \\
\hline
Rifky\_IndoBERT\_QA & 10.20 $\pm$ 1.37 & 32.62 $\pm$ 1.29 & 33.23 $\pm$ 1.62 \\
IndoLEM\_IndoBERT   &  5.59 $\pm$ 1.62 & 21.69 $\pm$ 2.37 & 22.25 $\pm$ 2.53 \\
Cahya\_IndoBERT     &  5.70 $\pm$ 0.94 & 23.75 $\pm$ 2.65 & 24.40 $\pm$ 3.04 \\
mT5 (Generative)    &  0.00 $\pm$ 0.00 &  6.60 $\pm$ 2.66 &  5.66 $\pm$ 2.30 \\
\hline
\end{tabular}}
\caption{$5$-fold cross-validation, folds stratified by source report. Mean $\pm$ standard deviation across folds.}
\label{tab:cv_results}
\end{table}

\subsection*{Learning Curve}
\begin{figure}[!tbp]
    \centering
    \includegraphics[width=0.85\columnwidth]{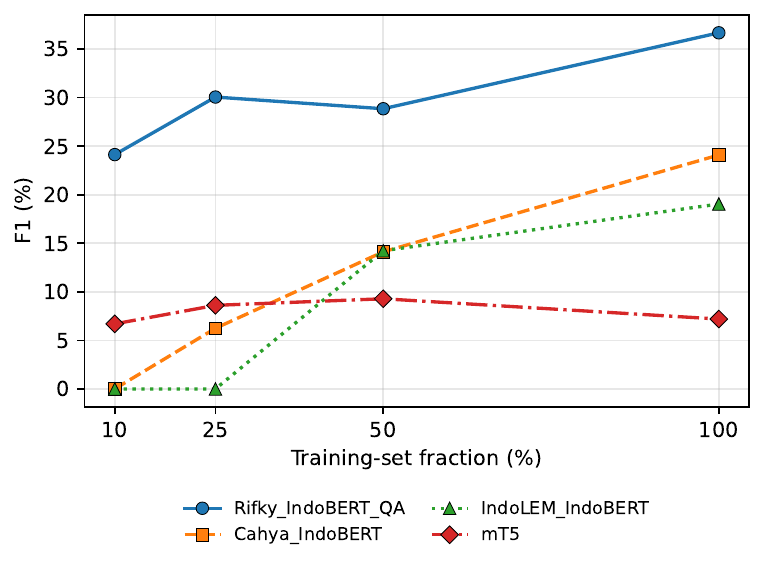}
    \caption{Learning curve: F1 on the held-out test set as a function of training-set fraction.}
    \label{fig:learning_curve}
\end{figure}

To quantify the marginal value of additional expert annotation, four of the fine-tuned models (Rifky\_IndoBERT\_QA, Cahya\_IndoBERT, IndoLEM\_IndoBERT, and mT5) are re-trained on $\{10, 25, 50, 100\}\%$ of the training split (validation and test splits held fixed). Figure~\ref{fig:learning_curve} plots F1 on the held-out test set. The strongest extractive baseline rises overall (with a small dip at $50\%$) and is still climbing at $100\%$, indicating additional annotation would likely yield gains; mT5 plateaus near zero across all fractions, indicating its bottleneck is architectural, not data.

\section{Comparison with Existing Medical QA Datasets}
\label{appx:comparison}

This appendix compares IndoNutriQA with the medical and Indonesian QA resources discussed in Section~\ref{sec:related}. Table~\ref{tab:positioning} summarises the positioning on the three properties the benchmark targets: Indonesian language, nutrition domain, and grounding in authentic clinical documentation. TyDi QA covers Indonesian but is not clinical, and emrQA is grounded in clinical records but is English and template-generated. No other resource in the table combines all three properties.

Table~\ref{tab:benchmark_comparison} extends this comparison to the full set of benchmarks, with language, domain, sample count, question and answer types, source, and average question and answer length. Three differences stand out. First, at $911$ pairs IndoNutriQA is the smallest resource listed, which is why the paper positions it as an evaluation benchmark rather than a training corpus (Section~\ref{sec:experiments}). Second, its answers are free text grounded in a clinical report. The exam-based resources (MedQA, MedMCQA, MedExQA) provide a selected option, and PubMedQA questions are answered yes, no or maybe, whereas emrQA answers are spans in an electronic health record. Third, its mean answer length of $70.66$ tokens is closer to long-answer resources such as MedQuAD ($197$ words) and MedExQA ($82.5$ words) than to the exam options ($2.7$--$20.6$ tokens or characters) or the emrQA spans ($18.7$ tokens). Length units are those reported by each source (tokens, words or characters), and tokenisation differs, so these comparisons are approximate.

\begin{table}[!tbp]
\centering
\footnotesize
\setlength{\tabcolsep}{3pt}
\begin{tabular}{lccc}
\toprule
Benchmark & Indo. & Nutr. & Clin.\ docs \\
\midrule
SQuAD        & --       & --       & -- \\
TyDi QA         & $\surd$  & --       & -- \\
PubMedQA        & --       & --       & -- \\
MedQA            & --       & --       & -- \\
MedMCQA          & --       & --       & -- \\
MedQuAD       & --       & --       & -- \\
emrQA          & --       & --       & $\surd$ \\
\midrule
\textbf{IndoNutriQA}                   & $\surd$  & $\surd$  & $\surd$ \\
\bottomrule
\end{tabular}
\caption{Positioning of IndoNutriQA against representative QA benchmarks. ``Clinical docs'' marks resources grounded in authentic clinical documentation rather than exam items or literature abstracts. The full nine-column comparison, including sample counts and question/answer lengths, is in Appendix~\ref{appx:comparison}.}
\label{tab:positioning}
\end{table}

\begin{table*}[!t]
\centering

\resizebox{\textwidth}{!}{
\begin{tabular}{l l l r l l l l l}
\hline
\makecell{\textbf{Dataset}}
& \makecell{\textbf{Language}}
& \makecell{\textbf{Domain}}
& \makecell{\textbf{\# Samples}}
& \makecell{\textbf{Q Type}}
& \makecell{\textbf{A Type}}
& \makecell{\textbf{Source}}
& \makecell{\textbf{Avg Q}\\\textbf{Length}}
& \makecell{\textbf{Avg A}\\\textbf{Length}} \\
\hline
MedQA (USMLE)~\cite{jin2021disease}     & EN & General Med   & 12{,}723   & MCQ        & Single     & Exam        & 116.6 tokens   & 3.5 tokens \\
MedQA (MCMLE)~\cite{jin2021disease}     & ZH & General Med   & 34{,}251   & MCQ        & Single     & Exam        & 45.7 chars     & 7.3 chars \\
MedQA (TWMLE)~\cite{jin2021disease}     & ZH & General Med   & 14{,}123   & MCQ        & Single     & Exam        & 61.0 chars     & 20.6 chars \\
MedMCQA~\cite{pal2022medmcqa}            & EN & General Med   & 194{,}000+ & MCQ        & Single     & Exam        & 12.71 tokens   & 2.70 tokens \\
PubMedQA (PQA-L)~\cite{jin2019pubmedqa}  & EN & Biomedical    & 1{,}000+   & Y/N/Maybe  & Short      & Literature  & 14.4 tokens    & 43.2 tokens \\
PubMedQA (PQA-U)~\cite{jin2019pubmedqa}  & EN & Biomedical    & 61{,}200+  & Y/N/Maybe  & Short      & Literature  & 15.0 tokens    & 45.9 tokens \\
PubMedQA (PQA-A)~\cite{jin2019pubmedqa}  & EN & Biomedical    & 211{,}300+ & Y/N/Maybe  & Short      & Literature  & 16.3 tokens    & 41.0 tokens \\
emrQA~\cite{pampari2018emrqa}            & EN & Clinical      & 400{,}000+ & Extractive & Span       & EHR         & 8.6 tokens     & 18.7 tokens \\
MedQuAD~\cite{abacha2019medquad}         & EN & General Med   & 47{,}457   & Open       & Long       & NIH Web     & 7 words        & 197 words \\
MedExQA~\cite{kim2024medexqa}            & EN & 5 Specialties & 3{,}500+   & MCQ        & Single+Exp & Exam        & 19.1 tokens    & 82.5 words \\
\midrule
IndoNutriQA                              & ID & Nutrition+Med & 911        & Diverse    & Contextual & Clin.\ Rep. & 8.32 tokens    & 70.66 tokens \\
\hline
\end{tabular}}
\caption{Combined overview and statistical comparison of medical QA datasets.}
\label{tab:benchmark_comparison}
\end{table*}

\section{RAGAS Metrics for the RAG Baseline}
\label{appx:ragas}

\subsection*{RAG Diagnostic Evaluation: full discussion}

The near-zero lexical numbers for the RAG baseline (Table~\ref{tab:model_performance}) cannot localize the failure to retrieval or generation. We compute the four RAGAS metrics \cite{es2024ragas} on the same test split, scored by GPT-4o-mini with \texttt{text-embedding-3-small} (Table~\ref{tab:ragas_metrics}, Appendix~\ref{appx:ragas}). In this configuration retrieval is the stronger component (context precision $73.43$, recall $35.98$, which is only moderate) while generation collapses (faithfulness $12.86$ over the $105$ examples on which it is defined, answer relevancy $0.24$), pointing to \texttt{mt5-base} rather than the retriever as the binding constraint here. The $32$ remaining examples are themselves evidence: RAGAS cannot extract a single atomic claim from those generated answers, so faithfulness is undefined: roughly one generated answer in four is too degenerate to score at all.

\noindent\textbf{Scope of this result.} We draw no conclusion about the effectiveness of RAG for Indonesian clinical QA. This is a diagnostic of one configuration (a single dense retriever over recursive chunks feeding an un-fine-tuned \texttt{mt5-base} generator), and it localises the binding constraint to the generator, not retrieval. Establishing what RAG can achieve here needs a stronger Indonesian generator with grounding-aware training and a sweep over retrievers and chunking; we leave both to future work.

Table~\ref{tab:ragas_metrics} reports the four RAGAS metrics for the RAG baseline on the test split, decomposing the near-zero lexical score into generator- and retriever-side components. The diagnosis (Section~\ref{subsec:ragas_evaluation}) is that, in this configuration, retrieval is the stronger component and \texttt{mt5-base} generation the weaker.

\begin{table}[!tbp]
\centering

\resizebox{0.8\columnwidth}{!}{
\begin{tabular}{lc}
\hline
\textbf{Metric} & \textbf{Score \% [95\% CI]} \\
\hline
\multicolumn{2}{l}{\textit{Generator-side}} \\
Faithfulness ($n=105$)      & 12.86 [--, --] \\
Answer relevancy            &  0.24 [ 0.00,  0.60] \\
\hline
\multicolumn{2}{l}{\textit{Retriever-side}} \\
Context precision           & 73.43 [66.85, 80.23] \\
Context recall              & 35.98 [29.61, 42.86] \\
\hline
\end{tabular}}
\caption{RAGAS metrics for the RAG baseline on the test split ($n = 137$). Scores in percent ($\times 100$) with bootstrap $95\%$ CIs. Faithfulness reported on the $105$ examples for which RAGAS could parse atomic claims; $32$ examples produced near-empty outputs on which faithfulness is undefined.}
\label{tab:ragas_metrics}
\end{table}

\section{Semantic and LLM-Judge Metric Details}
\label{appx:semantic}

\noindent\textbf{Metric normalisation conventions.} Exact match and token-F1 are computed after SQuAD-style normalisation (lowercasing, punctuation stripping, article removal, whitespace collapsing), whereas BLEU uses raw whitespace tokenisation and ROUGE uses \texttt{rouge\_score}'s internal tokenisation and stemming. BLEU is therefore punctuation-sensitive and reads systematically lower than F1 on the same predictions; because the convention is applied uniformly across systems, model rankings are unaffected. The SQuAD-derived normaliser also removes the English articles a, an, and the, which on Indonesian clinical text deletes the token A from nutrient names such as vitamin A; $67$ of the $911$ reference answers contain a standalone a. Since the same normaliser is applied to predictions and references alike F1 is essentially unaffected, but exact match cannot distinguish vitamin from vitamin A. Future releases of the evaluation script will drop the article-removal step.

Table~\ref{tab:semantic_metrics} reports per-model BERTScore F1, multilingual sentence-embedding cosine, and the four LLM-as-judge dimensions (factuality, completeness, safety, overall) with bootstrap $95\%$ CIs over $1{,}000$ resamples. BERTScore F1 and embedding cosine are reported as percentages; judge scores are on a $1$-$5$ Likert scale. Lexical F1 is reproduced from Table~\ref{tab:model_performance} for direct comparison.

\begin{table*}[!t]
\centering

\resizebox{\textwidth}{!}{
\begin{tabular}{l c c c c c c c}
\hline
\textbf{Model} & \textbf{Lexical F1} & \textbf{BERTScore F1} & \textbf{Embed.\ cos} & \textbf{Judge fact.} & \textbf{Judge compl.} & \textbf{Judge safety} & \textbf{Judge overall} \\
\hline
\multicolumn{8}{l}{\textit{Fine-tuned baselines}} \\
Rifky\_IndoBERT\_QA       & 36.34 & 45.94 [38.41, 54.07] & 45.27 [39.31, 51.74] & 3.14 [2.84, 3.45] & 2.84 [2.53, 3.13] & 4.89 [4.80, 4.96] & 3.62 [3.43, 3.83] \\
IndoLEM\_IndoBERT         & 23.78 & 35.42 [27.87, 42.57] & 36.23 [30.51, 41.84] & 2.61 [2.34, 2.91] & 2.33 [2.07, 2.60] & 4.85 [4.77, 4.92] & 3.26 [3.09, 3.45] \\
Cahya\_IndoBERT           & 16.56 & 25.78 [19.24, 32.26] & 28.61 [23.64, 33.79] & 2.46 [2.18, 2.75] & 2.19 [1.93, 2.45] & 4.85 [4.77, 4.92] & 3.17 [2.99, 3.36] \\
XLM-R-large               & 12.60 & 23.16 [16.74, 29.64] & 27.61 [23.02, 32.62] & 2.34 [2.05, 2.64] & 2.12 [1.88, 2.38] & 4.84 [4.75, 4.92] & 3.10 [2.92, 3.29] \\
mDeBERTa-v3-base          & 19.35 & 28.47 [21.53, 35.90] & 32.18 [26.71, 38.05] & 2.54 [2.28, 2.83] & 2.32 [2.07, 2.59] & 4.89 [4.82, 4.94] & 3.25 [3.08, 3.44] \\
mT5 (Generative)          &  7.84 & 79.11 [78.54, 79.70] & 18.19 [15.09, 21.40] & 1.08 [1.02, 1.15] & 1.04 [1.01, 1.08] & 4.62 [4.43, 4.80] & 2.25 [2.18, 2.32] \\
\hline
\multicolumn{8}{l}{\textit{Closed-source LLM}} \\
GPT-4o-mini (0-shot)              & 36.06 & 88.08 [87.43, 88.74] & 61.13 [58.11, 64.08] & 4.19 [4.01, 4.36] & \textbf{3.86} [3.71, 4.02] & 4.93 [4.89, 4.97] & 4.33 [4.22, 4.44] \\
GPT-4o-mini (5-shot)              & 43.07 & 89.40 [88.67, 90.17] & 66.51 [63.18, 69.77] & \textbf{4.36} [4.20, 4.52] & 3.80 [3.61, 3.96] & 4.96 [4.93, 4.99] & \textbf{4.37} [4.26, 4.47] \\
GPT-4o (0-shot)                   & 37.49 & 88.07 [87.24, 88.82] & 57.38 [52.96, 61.48] & 3.94 [3.69, 4.19] & 3.67 [3.43, 3.88] & 4.96 [4.91, 4.99] & 4.19 [4.03, 4.34] \\
GPT-4o (5-shot)                   & 50.33 & 90.62 [89.75, 91.51] & 69.66 [65.42, 74.14] & 4.21 [4.01, 4.41] & 3.58 [3.37, 3.80] & \textbf{4.98} [4.95, 5.00] & 4.26 [4.12, 4.38] \\
\hline
\multicolumn{8}{l}{\textit{Open-source multilingual LLM}} \\
Llama-3.1-8B (0-shot)             & 37.82 & 88.32 [87.61, 89.04] & 60.95 [57.01, 64.50] & 3.94 [3.72, 4.15] & 3.72 [3.51, 3.90] & 4.87 [4.81, 4.93] & 4.18 [4.03, 4.31] \\
Llama-3.1-8B (5-shot)             & \textbf{55.55} & 91.09 [90.20, 91.94] & \textbf{74.37} [70.58, 78.24] & 3.95 [3.72, 4.17] & 3.43 [3.20, 3.65] & 4.88 [4.82, 4.93] & 4.09 [3.93, 4.24] \\
Qwen2.5-7B (0-shot)               & 31.03 & 87.35 [86.69, 87.99] & 58.28 [55.21, 61.30] & 3.73 [3.53, 3.93] & 3.62 [3.47, 3.77] & 4.82 [4.74, 4.88] & 4.06 [3.94, 4.18] \\
Qwen2.5-7B (5-shot)               & 42.62 & 89.48 [88.65, 90.26] & 66.50 [62.98, 70.08] & 4.12 [3.93, 4.28] & 3.64 [3.46, 3.80] & 4.91 [4.85, 4.96] & 4.22 [4.10, 4.33] \\
\hline
\multicolumn{8}{l}{\textit{Indonesian-tuned LLM}} \\
Sahabat-AI-Llama3-8B (0-shot)     & 26.78 & 85.89 [85.24, 86.56] & 64.09 [61.15, 67.06] & 4.05 [3.84, 4.26] & 3.58 [3.36, 3.81] & 4.92 [4.88, 4.96] & 4.18 [4.04, 4.32] \\
Sahabat-AI-Llama3-8B (5-shot)     & 43.32 & 89.97 [88.97, 90.99] & 71.69 [67.99, 75.59] & 3.54 [3.27, 3.82] & 3.01 [2.77, 3.25] & 4.67 [4.54, 4.78] & 3.74 [3.56, 3.93] \\
Merak-7B-v4 (0-shot)              & 32.44 & 88.00 [87.35, 88.69] & 61.48 [58.01, 64.88] & 3.69 [3.45, 3.91] & 3.43 [3.23, 3.62] & 4.74 [4.64, 4.82] & 3.95 [3.80, 4.10] \\
Merak-7B-v4 (5-shot)              & 41.83 & 89.95 [89.17, 90.81] & 69.43 [65.98, 73.06] & 3.82 [3.61, 4.04] & 3.18 [2.99, 3.37] & 4.85 [4.77, 4.91] & 3.95 [3.82, 4.09] \\
\hline
\multicolumn{8}{l}{\textit{English-medical LLM}} \\
BioMistral-7B (0-shot)            & 29.83 & 74.76 [69.61, 79.76] & 53.05 [48.38, 57.77] & 2.84 [2.57, 3.13] & 2.40 [2.18, 2.65] & 4.90 [4.82, 4.96] & 3.38 [3.22, 3.55] \\
BioMistral-7B (5-shot)            & 40.08 & 89.38 [88.55, 90.31] & 66.14 [62.14, 70.35] & 3.28 [3.03, 3.55] & 2.66 [2.46, 2.88] & 4.74 [4.64, 4.82] & 3.56 [3.40, 3.73] \\
MedGemma-4B-IT (0-shot)           & 42.60 & 88.79 [88.04, 89.57] & 63.50 [60.08, 66.87] & 3.97 [3.77, 4.17] & 3.66 [3.46, 3.83] & 4.92 [4.86, 4.96] & 4.18 [4.05, 4.31] \\
MedGemma-4B-IT (5-shot)           & 54.28 & \textbf{91.12} [90.20, 92.03] & 73.56 [69.91, 77.52] & 3.82 [3.59, 4.05] & 3.37 [3.15, 3.61] & 4.88 [4.81, 4.94] & 4.02 [3.87, 4.18] \\
\hline
\end{tabular}}
\caption{Semantic and LLM-judge metrics on the IndoNutriQA test split ($n = 137$), with bootstrap $95\%$ CIs.}
\label{tab:semantic_metrics}
\end{table*}

\subsection*{Second-Judge Cross-Validation (Claude Haiku 4.5)}

To verify that the LLM-as-judge scores are not artefacts of a single judge family, we re-score the same $137$-example test split with Claude Haiku 4.5 (Anthropic) under identical prompts and decoding settings. Per-example inter-judge agreement between GPT-4o-mini and Claude Haiku 4.5 across the three rubric dimensions ($n = 2{,}879$ paired judgements, pooled across all $22$ model rows) is summarised in Table~\ref{tab:judge_agreement}; Claude-judge model-level scores are reported in Table~\ref{tab:semantic_metrics_claude}.

\begin{table}[!htbp]
\centering
\resizebox{\columnwidth}{!}{
\begin{tabular}{lccc}
\hline
\textbf{Dimension} & \textbf{Quadratic $\kappa$} & \textbf{Within-1 pt.} & \textbf{Mean abs.\ diff.} \\
\hline
Factuality            & 0.584 & 83.3\% & 0.77 \\
Completeness          & 0.711 & 84.6\% & 0.71 \\
Safety                & 0.251 & 80.7\% & 0.66 \\
\hline
Pooled (all dims)     & 0.652 & 82.8\% & 0.71 \\
\hline
\end{tabular}}
\caption{Inter-judge agreement between GPT-4o-mini and Claude Haiku 4.5 across the three rubric dimensions. Quadratic-weighted Cohen's $\kappa$ over the $1$--$5$ Likert ordinal scale; ``within-1'' = fraction of paired judgements that differ by $\le 1$ Likert point.}
\label{tab:judge_agreement}
\end{table}

Pooled $\kappa = 0.652$ indicates substantial agreement under the Landis--Koch rubric, with the two judges within $1$ Likert point on $82.8\%$ of paired judgements. The dimension-level breakdown is intuitive: completeness has the highest agreement ($\kappa = 0.711$) because both judges look for missing reference content; factuality is moderate ($\kappa = 0.584$); safety is the lowest ($\kappa = 0.251$) because both judges concentrate scores near the ceiling (mean $\approx 4.8$/$5$ across all models), which compresses the kappa denominator; the within-1-point agreement on safety remains high ($80.7\%$). Importantly, the top-tier ranking on the overall-rubric score is preserved across both judges: GPT-4o (5-shot) wins safety under both, and GPT-4o-mini / GPT-4o / Llama-3.1-8B / MedGemma form a tight top tier on the overall score under both judges. The full Claude-judge model-level scores are listed in Table~\ref{tab:semantic_metrics_claude}.

\begin{table*}[!t]
\centering

\small
\begin{tabular}{l c c c c}
\hline
\textbf{Model} & \textbf{Judge fact.} & \textbf{Judge compl.} & \textbf{Judge safety} & \textbf{Judge overall} \\
\hline
\multicolumn{5}{l}{\textit{Fine-tuned baselines}} \\
Rifky\_IndoBERT\_QA              & 3.98 [3.68, 4.29]            & 2.93 [2.60, 3.26]            & 4.28 [4.02, 4.53]            & 3.73 [3.46, 3.98]            \\
IndoLEM\_IndoBERT                & 3.80 [3.49, 4.10]            & 2.35 [2.05, 2.66]            & 4.17 [3.93, 4.41]            & 3.44 [3.20, 3.68]            \\
Cahya\_IndoBERT                  & 3.85 [3.52, 4.18]            & 2.23 [1.90, 2.57]            & 4.21 [3.93, 4.45]            & 3.43 [3.14, 3.69]            \\
XLM-R-large                      & 3.81 [3.51, 4.12]            & 1.96 [1.69, 2.29]            & 4.12 [3.86, 4.38]            & 3.30 [3.07, 3.55]            \\
mDeBERTa-v3-base                 & 3.87 [3.58, 4.17]            & 2.20 [1.88, 2.54]            & 4.18 [3.93, 4.42]            & 3.42 [3.18, 3.65]            \\
mT5 (Generative)                 & 1.60 [1.42, 1.78]            & 1.06 [1.02, 1.09]            & 3.03 [2.81, 3.26]            & 1.90 [1.78, 2.01]            \\
\hline
\multicolumn{5}{l}{\textit{Closed-source LLM}} \\
GPT-4o-mini (0-shot)             & 3.99 [3.82, 4.16]            & 3.23 [3.04, 3.42]            & 4.34 [4.18, 4.48]            & 3.86 [3.70, 4.00]            \\
GPT-4o-mini (5-shot)             & 4.15 [3.98, 4.31]            & \textbf{3.32} [3.09, 3.54]   & 4.50 [4.37, 4.64]            & 3.99 [3.83, 4.14]            \\
GPT-4o (0-shot)                  & 4.09 [3.89, 4.28]            & 3.10 [2.86, 3.33]            & 4.53 [4.39, 4.65]            & 3.91 [3.76, 4.05]            \\
GPT-4o (5-shot)                  & \textbf{4.39} [4.22, 4.56]   & 3.29 [3.06, 3.51]            & \textbf{4.75} [4.64, 4.85]   & \textbf{4.15} [4.00, 4.28]   \\
\hline
\multicolumn{5}{l}{\textit{Open-source multilingual LLM}} \\
Llama-3.1-8B (0-shot)            & 3.92 [3.73, 4.11]            & 3.09 [2.87, 3.31]            & 4.30 [4.14, 4.44]            & 3.77 [3.60, 3.93]            \\
Llama-3.1-8B (5-shot)            & 4.12 [3.91, 4.30]            & 3.20 [2.97, 3.44]            & 4.42 [4.26, 4.57]            & 3.91 [3.74, 4.08]            \\
Qwen2.5-7B (0-shot)              & 3.69 [3.51, 3.88]            & 2.96 [2.75, 3.18]            & 4.15 [4.00, 4.31]            & 3.60 [3.45, 3.76]            \\
Qwen2.5-7B (5-shot)              & 4.17 [3.99, 4.34]            & 3.16 [2.93, 3.38]            & 4.53 [4.38, 4.67]            & 3.95 [3.80, 4.10]            \\
\hline
\multicolumn{5}{l}{\textit{Indonesian-tuned LLM}} \\
Sahabat-AI-Llama3-8B (0-shot)    & 3.94 [3.73, 4.13]            & 2.80 [2.61, 3.00]            & 4.26 [4.09, 4.42]            & 3.66 [3.51, 3.82]            \\
Sahabat-AI-Llama3-8B (5-shot)    & 3.74 [3.50, 3.96]            & 2.85 [2.58, 3.12]            & 4.16 [3.96, 4.35]            & 3.58 [3.37, 3.78]            \\
Merak-7B-v4 (0-shot)             & 3.66 [3.45, 3.87]            & 2.69 [2.47, 2.92]            & 3.99 [3.77, 4.18]            & 3.45 [3.26, 3.63]            \\
Merak-7B-v4 (5-shot)             & 3.88 [3.67, 4.08]            & 2.77 [2.55, 3.00]            & 4.23 [4.04, 4.40]            & 3.62 [3.45, 3.80]            \\
\hline
\multicolumn{5}{l}{\textit{English-medical LLM}} \\
BioMistral-7B (0-shot)           & 3.68 [3.41, 3.93]            & 2.20 [1.98, 2.43]            & 4.11 [3.89, 4.31]            & 3.33 [3.14, 3.53]            \\
BioMistral-7B (5-shot)           & 3.65 [3.42, 3.87]            & 2.63 [2.39, 2.87]            & 3.96 [3.76, 4.16]            & 3.41 [3.21, 3.61]            \\
MedGemma-4B-IT (0-shot)          & 3.95 [3.75, 4.14]            & 3.04 [2.83, 3.26]            & 4.38 [4.23, 4.51]            & 3.79 [3.64, 3.95]            \\
MedGemma-4B-IT (5-shot)          & 4.00 [3.80, 4.20]            & 3.09 [2.85, 3.32]            & 4.29 [4.12, 4.47]            & 3.79 [3.62, 3.96]            \\
\hline
\end{tabular}
\caption{Claude Haiku 4.5 LLM-as-judge scores (the second-judge cross-validation of Table~\ref{tab:semantic_metrics}). Scale: $1$--$5$ Likert with bootstrap $95\%$ CIs over $1{,}000$ resamples. Bold marks the strongest model per dimension.}
\label{tab:semantic_metrics_claude}
\end{table*}

\subsection*{Judge Self-Preference Analysis}
\label{appx:judge_bias}

GPT-4o-mini serves as the primary judge while GPT-4o-mini and GPT-4o also appear as evaluated systems. We therefore test directly for self-preference. A raw comparison of the judge's scores on OpenAI systems against its scores on other systems would confound self-preference with the fact that the OpenAI systems are genuinely stronger. We instead use a difference-in-differences design. For each of the $22$ evaluated systems we compute the mean overall-rubric score under GPT-4o-mini minus the mean under Claude Haiku $4.5$, restricted to the examples both judges scored, so the per-system judge gap is a paired quantity from which true system quality, visible to both judges, has been differenced out. We then ask whether that gap is larger for systems sharing the judge's provider than for the rest. Claude Haiku $4.5$ is a valid control because no evaluated system comes from its provider, so it has no self to prefer.

Table~\ref{tab:judge_bias} reports the result. GPT-4o-mini is uniformly more lenient than Claude Haiku $4.5$ ($+0.186$ rubric points on non-OpenAI systems alone), but the additional leniency it extends to its own provider is $+0.125$ points on the overall score and is not statistically distinguishable from zero (permutation $p = 0.34$). The per-dimension breakdown argues further against a systematic mechanism: the estimated effect is positive on factuality ($+0.456$, $p = 0.103$) but negative on safety ($-0.259$, $p = 0.092$), i.e.\ the judge is if anything harsher on its own provider's safety. A genuine self-preference bias would push consistently in one direction. Across the eight tests reported, none is significant at a Bonferroni-corrected threshold ($\alpha = 0.00625$).

Two caveats. First, with only four same-provider systems the test is underpowered; we can report that no self-preference effect is detectable, not that none exists. Second, the design detects provider-level favouritism, not a stylistic prior that both judges might share; the expert-rater calibration of Appendix~\ref{appx:rd_judge} addresses the latter by anchoring both judges to human clinical judgement.

\subsection*{Expert-Rater Calibration of the LLM Judge}
\label{appx:rd_judge}

To anchor the LLM-as-judge rubric to human clinical judgement, three expert raters (one Registered Dietitian, two nutritionists) independently re-scored a stratified sample of model outputs. We drew $50$ (test example, system output) units stratified by question category (factual / inferential / recommendation) crossed with system quality tier, so that the sample spans the full answer-quality range; the $50$ units cover $50$ distinct test examples and all $20$ source reports in the test split, and were produced by $20$ of the $22$ evaluated systems. Each rater scored every unit on the same three-dimension, $1$--$5$ rubric the LLM judge uses (factuality, completeness, clinical safety), blind to system identity and to the judges' scores, working from the question, the reference answer, the candidate answer, and the patient's diagnosis and anthropometric assessment. Stratification was on question category and aggregate system quality, never on the per-unit judge score, so the sample does not condition on the quantity being estimated.

Table~\ref{tab:rd_agreement} (top) reports quadratic-weighted Cohen's $\kappa$ for the reference rater pair and Krippendorff's ordinal $\alpha$ across all three. Factuality ($\kappa = 0.780$, $\alpha = 0.592$) and completeness ($\kappa = 0.752$, $\alpha = 0.712$) reach substantial agreement. Clinical safety does not ($\kappa = 0.160$, $\alpha = 0.250$): the three raters' mean safety scores are $3.88$, $4.66$, and $2.94$, a spread of $1.72$ rubric points between the most and least lenient. Safety, as operationalised by this rubric, is therefore not a dimension on which raters themselves converge, and any judge-vs-human safety comparison inherits that instability.

Table~\ref{tab:rd_agreement} (middle) compares each judge against the rater consensus (per-dimension mean, rounded half up). The key comparison is vertical: on factuality, GPT-4o-mini agrees with the rater consensus at $\kappa = 0.705$ against an inter-rater ceiling of $0.780$, and on completeness at $0.623$ against a ceiling of $0.752$; the confidence intervals overlap substantially in both cases. On these two dimensions the judge is therefore about as close to a rater as raters are to each other, which is the strongest validation available for a rubric of this kind. Agreement is highest on the recommendation questions (factuality $\kappa = 0.918$) and weakest on the short-span factual ones ($0.478$), mirroring the difficulty ordering in Section~\ref{subsec:stratified_analysis}.

Safety is the exception on both axes. Judge-vs-consensus $\kappa$ is $0.133$ for GPT-4o-mini and $0.480$ for Claude Haiku~4.5, and the leniency column shows why: GPT-4o-mini scores safety $+1.14$ rubric points above the rater consensus (mean $4.92$ against $3.78$), essentially saturating the scale, and Claude $+0.58$. Both judges systematically overstate the clinical safety of model output. Two further asymmetries are worth recording. GPT-4o-mini is not uniformly lenient: on the factual questions it is stricter than the raters ($-0.82$ on factuality). Claude Haiku~4.5 systematically under-scores completeness ($-0.78$ pooled). We therefore recommend that the safety dimension of this rubric be treated as unvalidated, that Claude Haiku~4.5 be preferred where a safety signal is needed ($\kappa = 0.480$ vs $0.133$), and that neither judge's absolute safety score be read as a clinical guarantee.

Raters could additionally flag units whose reference answer they judged incorrect or contestable. Three of $50$ units were flagged ($6\%$), one of them by two raters independently; the flagged cases concern macronutrient percentage distributions that do not sum to $100\%$, a mid-upper-arm-circumference cut-off, and a neonatal protein prescription. This is a direct estimate of gold-answer error in the test split and is consistent with the single-annotator caveat recorded in the Limitations; the flagged items are released with the benchmark so that users can exclude or re-adjudicate them.

\begin{table}[!tbp]
\centering
\small
\begin{tabular}{lccc}
\toprule
 & Fact. & Compl. & Safety \\
\midrule
\multicolumn{4}{l}{\textit{Inter-rater ($n=50$)}} \\
Quadratic $\kappa$      & $0.780$ & $0.752$ & $0.160$ \\
Krippendorff $\alpha$   & $0.592$ & $0.712$ & $0.250$ \\
\midrule
\multicolumn{4}{l}{\textit{Judge vs rater consensus (quadratic $\kappa$)}} \\
GPT-4o-mini             & $0.705$ & $0.623$ & $0.133$ \\
Claude Haiku 4.5        & $0.674$ & $0.585$ & $0.480$ \\
\midrule
\multicolumn{4}{l}{\textit{Judge leniency (judge $-$ consensus, rubric points)}} \\
GPT-4o-mini             & $-0.14$ & $-0.20$ & $+1.14$ \\
Claude Haiku 4.5        & $0.00$  & $-0.78$ & $+0.58$ \\
\bottomrule
\end{tabular}
\caption{Expert-rater calibration of the LLM-as-judge rubric on $50$ stratified units. Inter-rater $\kappa$ is the reference pair; Krippendorff's ordinal $\alpha$ pools all three raters. Judge agreement on factuality and completeness approaches the inter-rater ceiling, whereas safety agreement is low and both judges score safety substantially above the rater consensus.}
\label{tab:rd_agreement}
\end{table}

Finally, the choice of judge does not materially reorder the leaderboard (Spearman $\rho = 0.916$ between the two judges' system rankings), and the substantive conclusion, that closed-source models lead the clinical rubric while open-source models lead the similarity axes, holds under both judges. The single rank that does move is the top position itself: GPT-4o-mini leads under its own family's judge ($4.37$) whereas GPT-4o leads under Claude ($4.15$). We therefore report the rubric leader as a top tier rather than a single winner.

\begin{table}[!tbp]
\centering
\small
\begin{tabular}{lrrrr}
\toprule
 & \multicolumn{2}{c}{Mean judge gap} & & \\
\cmidrule(lr){2-3}
Dimension & Same prov. & Other & DiD & $p$ \\
 & ($n{=}4$) & ($n{=}18$) & & \\
\midrule
Factuality    & $+0.018$ & $-0.438$ & $+0.456$ & $0.103$ \\
Completeness  & $+0.491$ & $+0.313$ & $+0.178$ & $0.225$ \\
Safety        & $+0.425$ & $+0.684$ & $-0.259$ & $0.092$ \\
\midrule
Overall       & $+0.311$ & $+0.186$ & $+0.125$ & $0.340$ \\
\bottomrule
\end{tabular}
\caption{Judge self-preference test. ``Judge gap'' is the per-system mean score under GPT-4o-mini minus the mean under Claude Haiku $4.5$ on commonly scored examples; positive values mean GPT-4o-mini is more lenient. ``Same prov.'' groups the four systems sharing the judge's provider (GPT-4o-mini and GPT-4o, zero- and five-shot). DiD is the difference-in-differences estimate of self-preference; $p$ is a two-sided permutation test over $20{,}000$ relabellings. No dimension shows a significant effect, and the factuality and safety estimates carry opposite signs.}
\label{tab:judge_bias}
\end{table}

\section{Human Error Analysis Details}
\label{appx:human_error}

The nine categories of Table~\ref{tab:human_errors} were obtained by open-coding the $53$ free-text notes the three expert raters left during the calibration study of Appendix~\ref{appx:rd_judge}, then re-applying the finalised codebook to every note from scratch. No taxonomy was shown to the annotators, so error identification is theirs and unprompted; the coding is ours. A note may carry several codes (a rater often names an omission and a format violation in one sentence), so codes sum to more than $53$.

Table~\ref{tab:human_errors} reports, for each category, the number and percentage of the $50$ units in which it occurs; a unit may carry several categories, so the percentages sum to more than $100\%$. ``Mean rater score'' is the raters' own $1$--$5$ rating (factuality, completeness, safety) on the units carrying that category. Two categories have no counterpart in the LLM-judged taxonomy of Table~\ref{tab:error_taxonomy}: NCP/PES violation, and unsupported addition, for which the judge's hallucination label did not flag the units.

Notes were optional, and the rate differs sharply by rater ($36$, $9$, and $8$ notes), so each count is a lower bound on prevalence and not an error rate; the category ordering is more reliable than the absolute percentages. The $50$ units span $20$ systems, so the analysis characterises failure modes of the evaluated pool rather than of any one system. Because the raters saw the patient's diagnosis and anthropometric assessment while the judge saw only reference and candidate, part of the judge-vs-human gap on unsupported content is an information gap rather than a reasoning gap, which is itself an argument for giving judges the record.

Numeric errors concentrate on the recommendation questions, which carry prescription arithmetic. And degenerate output (repetition, or report boilerplate and table markup leaking into the answer) still draws high rubric scores ($4.40/4.20/4.40$): the raters noted the defect in prose but did not treat it as a clinical error, which is a rubric-design gap rather than a model behaviour.

\begin{table}[!tbp]
\centering\footnotesize
\setlength{\tabcolsep}{4pt}
\resizebox{0.8\columnwidth}{!}{
\begin{tabular}{@{}lrrccc@{}}
\toprule
 & \multicolumn{2}{c}{Units} & \multicolumn{3}{c}{Mean rater score} \\
\cmidrule(lr){2-3}\cmidrule(lr){4-6}
Human error category & $n$ & \% & Fact. & Compl. & Safe. \\
\midrule
Omission                & $18$ & $36$ & $3.32$ & $2.58$ & $2.53$ \\
Non-answer / truncated   & $8$  & $16$ & $1.75$ & $1.50$ & $1.50$ \\
NCP/PES violation        & $6$  & $12$ & $3.20$ & $2.60$ & $2.40$ \\
Reference itself wrong   & $5$  & $10$ & $2.83$ & $3.67$ & $2.50$ \\
Unsupported addition     & $5$  & $10$ & $4.17$ & $3.83$ & $4.00$ \\
Degenerate text          & $4$  & $8$  & $4.40$ & $4.20$ & $4.40$ \\
Contradicts patient data & $3$  & $6$  & $2.00$ & $2.00$ & $1.80$ \\
Numeric error            & $3$  & $6$  & $2.00$ & $3.00$ & $1.75$ \\
Non-standard language    & $3$  & $6$  & $4.00$ & $4.00$ & $3.67$ \\
\bottomrule
\end{tabular}}
\caption{Human error analysis: $53$ free-text notes from three expert raters over $37$ of $50$ audited units, coded into nine categories. \% is of the $50$ units; counts are lower bounds. Details are in Appendix~\ref{appx:human_error}.}
\label{tab:human_errors}
\end{table}

\begin{figure*}[!t]
    \centering
    \includegraphics[width=\textwidth]{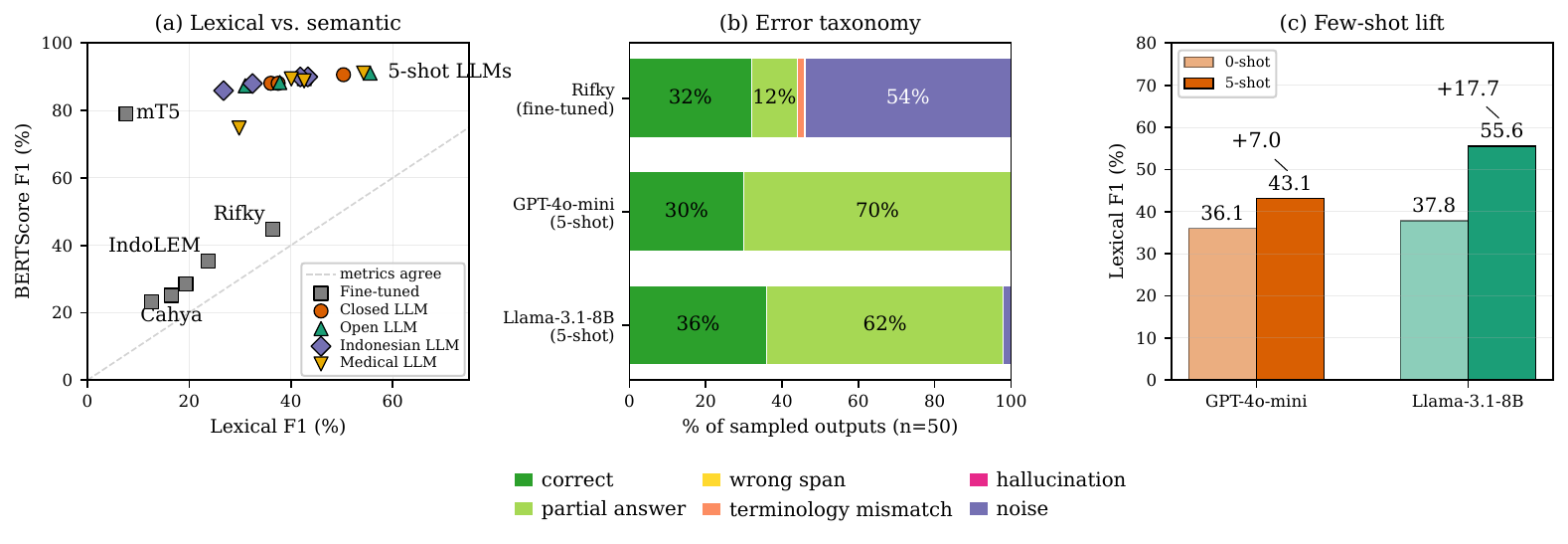}
    \caption{Headline findings on the IndoNutriQA test split ($n = 137$). \textbf{(a)} Lexical vs.\ BERTScore F1, markers shaped by model family: LLMs cluster above the ``metrics agree'' diagonal and mT5 falls far off it (high BERTScore from generic-token alignment, near-zero lexical), so no single metric suffices. \textbf{(b)} LLM-judge error taxonomy over $50$ sampled outputs per model (three representative rows): the extractive baseline is dominated by noise ([CLS]-token collapse), the LLMs by partial answer, and the judge's hallucination label is unreliable, as the human error analysis of Section~\ref{subsec:human_error} shows. \textbf{(c)} Five-shot prompting lifts Llama-3.1-8B by $+17.7$ F1 versus $+7.0$ for GPT-4o-mini, consistent with demonstrations mainly calibrating answer style.}
    \label{fig:main_results}
\end{figure*}

\section{Stratified Performance Tables}
\label{appx:stratified}

Figure~\ref{fig:main_results} summarizes the headline findings of the experiments: the gap between lexical and semantic scores (panel a), the error taxonomy of the LLM judge (panel b), and the effect of five-shot prompting (panel c). The tables that follow give the stratified results in detail.

\subsection*{Question-Category Breakdown}

Table~\ref{tab:question_breakdown} reports per-model F1 broken down by the question categories defined in Appendix~\ref{subsec:question_types} (factual, inferential, and recommendation; the comparative category is absent because the standardized questions each target a single time point per patient). This stratification isolates which reasoning types drive the aggregate scores in Section~\ref{subsec:model_performance}: recommendation and inferential questions, whose answers are longer and more structured, generally attract higher F1 than the shorter factual questions, indicating that headline performance is not uniform across reasoning complexity.

\begin{table*}[!t]
\centering

\small
\resizebox{0.75\textwidth}{!}{
\begin{tabular}{l c c c}
\hline
\textbf{Model} & \textbf{Factual ($n{=}38$)} & \textbf{Inferential ($n{=}39$)} & \textbf{Recommendation ($n{=}60$)} \\
\hline
\multicolumn{4}{l}{\textit{Fine-tuned baselines}} \\
Rifky\_IndoBERT\_QA       & 12.78 & 29.80 & 55.51 \\
IndoLEM\_IndoBERT         &  5.07 & 17.18 & 39.92 \\
Cahya\_IndoBERT           &  2.55 &  3.55 & 33.89 \\
XLM-R-large               &  2.09 &  4.43 & 24.56 \\
mDeBERTa-v3-base          &  3.07 &  6.05 & 38.30 \\
mT5 (Generative)          &  5.19 & 12.10 &  6.75 \\
\hline
\multicolumn{4}{l}{\textit{Closed-source LLM}} \\
GPT-4o-mini (0-shot)              & 21.83 & 40.46 & 42.20 \\
GPT-4o-mini (5-shot)              & 31.24 & 42.82 & 50.71 \\
GPT-4o (0-shot)                   & 18.47 & 51.91 & 40.17 \\
GPT-4o (5-shot)                   & 25.88 & \textbf{62.39} & 57.97 \\
\hline
\multicolumn{4}{l}{\textit{Open-source multilingual LLM}} \\
Llama-3.1-8B (0-shot)             & 21.14 & 43.31 & 44.82 \\
Llama-3.1-8B (5-shot)             & 29.64 & 59.55 & \textbf{69.36} \\
Qwen2.5-7B (0-shot)               & 18.46 & 33.98 & 37.07 \\
Qwen2.5-7B (5-shot)               & 29.23 & 52.12 & 44.91 \\
\hline
\multicolumn{4}{l}{\textit{Indonesian-tuned LLM}} \\
Sahabat-AI-Llama3-8B (0-shot)     & 19.91 & 29.32 & 29.48 \\
Sahabat-AI-Llama3-8B (5-shot)     & 21.38 & 49.61 & 53.13 \\
Merak-7B-v4 (0-shot)              & 22.55 & 38.33 & 34.88 \\
Merak-7B-v4 (5-shot)              & 28.96 & 47.12 & 46.55 \\
\hline
\multicolumn{4}{l}{\textit{English-medical LLM}} \\
BioMistral-7B (0-shot)            & 14.58 & 30.41 & 39.10 \\
BioMistral-7B (5-shot)            & 32.30 & 32.24 & 50.10 \\
MedGemma-4B-IT (0-shot)           & 23.90 & 52.61 & 47.92 \\
MedGemma-4B-IT (5-shot)           & \textbf{33.39} & 61.52 & 62.80 \\
\hline
\end{tabular}}
\caption{F1 by question category on the IndoNutriQA test split. Categories from Appendix~\ref{sec:appendix}: factual ($n = 38$), inferential ($n = 39$), recommendation ($n = 60$). The fourth category proposed in Appendix~\ref{subsec:question_types} (comparative) is not represented because the standardized questions all target a single time point per patient. Bold marks the strongest model per category.}
\label{tab:question_breakdown}
\end{table*}

\subsection*{Disease-Category Breakdown}

\begin{figure*}[!htbp]
    \centering
    \includegraphics[width=0.7\textwidth]{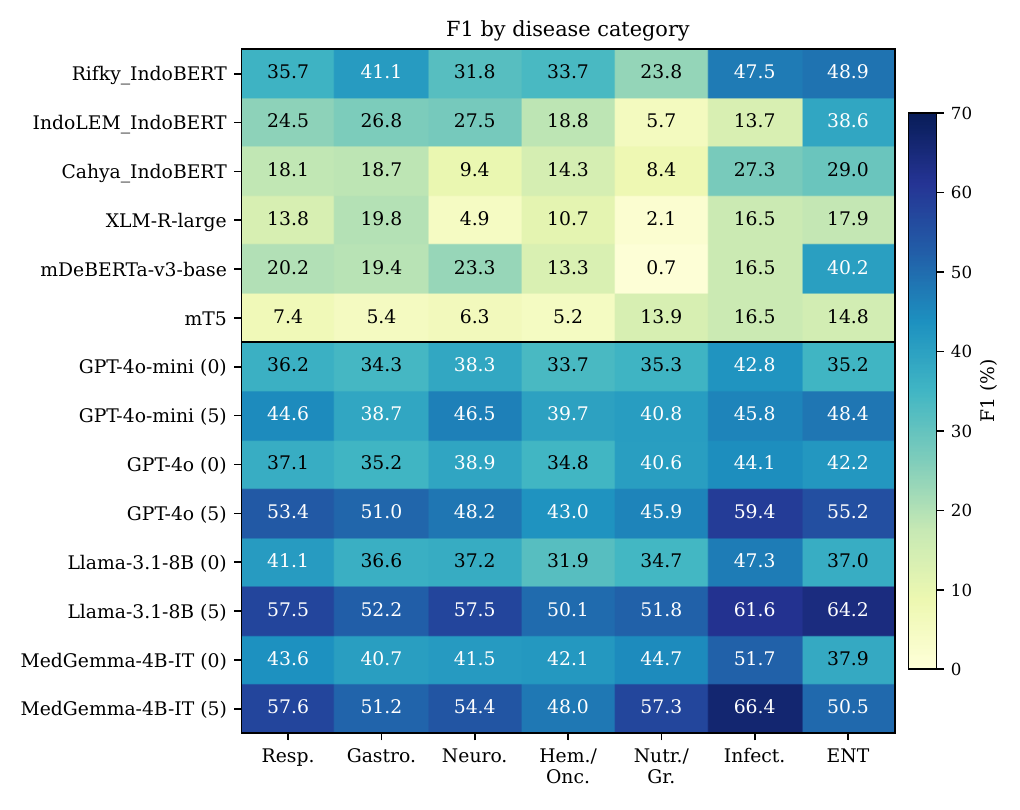}
    \caption{F1 by disease category for each evaluated model. Rows banded by family (fine-tuned baselines above the solid line, LLMs below); cells colored on a common scale. The respiratory column (the largest category, $28.33\%$ of source reports) is not systematically darker than the columns for categories with far fewer test-split reports (ENT, infectious, nutrition/growth), supporting the Section~\ref{subsec:domain_coverage} claim that the dataset imbalance does not translate to biased generalization.}
    \label{fig:disease_heatmap}
\end{figure*}

Figure~\ref{fig:disease_heatmap} visualizes the disease-category F1 matrix as a heatmap, making the absence of a systematic advantage for the largest category (respiratory) visible at a glance. Table~\ref{tab:disease_breakdown} gives the full numeric values.

\begin{table*}[!t]
\centering

\small
\begin{tabular}{l c c c c c c c}
\hline
\textbf{Model}
& \makecell{Resp.\\($n{=}42$)}
& \makecell{Gastro.\\($n{=}27$)}
& \makecell{Neuro.\\($n{=}25$)}
& \makecell{Hem./Onc.\\($n{=}22$)}
& \makecell{Nutr./Gr.\\($n{=}7$)}
& \makecell{Infect.\\($n{=}7$)}
& \makecell{ENT\\($n{=}7$)} \\
\hline
\multicolumn{8}{l}{\textit{Fine-tuned baselines}} \\
Rifky\_IndoBERT\_QA & 35.71 & 41.96 & 31.83 & 34.23 & 23.77 & 41.15 & 48.91 \\
IndoLEM\_IndoBERT   & 24.51 & 26.76 & 27.48 & 18.78 &  5.71 & 13.65 & 38.57 \\
Cahya\_IndoBERT     & 18.37 & 18.72 &  9.35 & 13.89 &  8.37 & 27.27 & 29.00 \\
XLM-R-large         & 13.77 & 19.83 &  4.87 & 10.67 &  2.09 & 16.54 & 17.85 \\
mDeBERTa-v3-base    & 20.22 & 19.40 & 23.29 & 13.32 &  0.68 & 16.53 & 40.24 \\
mT5 (Generative)    &  7.36 &  6.53 &  6.33 &  5.19 & 13.86 & 16.48 & 14.83 \\
\hline
\multicolumn{8}{l}{\textit{Closed-source LLM}} \\
GPT-4o-mini (0-shot)             & 36.20 & 34.35 & 38.28 & 33.73 & 35.34 & 42.75 & 35.16 \\
GPT-4o-mini (5-shot)             & 44.57 & 38.73 & 46.53 & 39.74 & 40.76 & 45.80 & 48.36 \\
GPT-4o (0-shot)                  & 37.10 & 35.20 & 38.93 & 34.83 & 40.55 & 44.13 & 42.23 \\
GPT-4o (5-shot)                  & 53.40 & 51.03 & 48.19 & 42.98 & 45.91 & 59.41 & 55.24 \\
\hline
\multicolumn{8}{l}{\textit{Open-source multilingual LLM}} \\
Llama-3.1-8B (0-shot)            & 41.14 & 36.58 & 37.23 & 31.95 & 34.71 & 47.28 & 36.95 \\
Llama-3.1-8B (5-shot)            & 57.55 & \textbf{52.23} & \textbf{57.54} & \textbf{50.08} & 51.83 & 61.56 & \textbf{64.17} \\
Qwen2.5-7B (0-shot)              & 30.49 & 29.90 & 31.92 & 31.18 & 35.21 & 30.98 & 30.83 \\
Qwen2.5-7B (5-shot)              & 44.23 & 44.85 & 40.72 & 33.29 & 41.95 & 41.50 & 62.17 \\
\hline
\multicolumn{8}{l}{\textit{Indonesian-tuned LLM}} \\
Sahabat-AI-Llama3-8B (0-shot)    & 27.88 & 26.59 & 27.02 & 25.18 & 29.35 & 23.57 & 25.61 \\
Sahabat-AI-Llama3-8B (5-shot)    & 44.64 & 39.63 & 48.81 & 43.22 & 40.86 & 38.05 & 38.12 \\
Merak-7B-v4 (0-shot)             & 31.57 & 33.67 & 33.31 & 30.91 & 33.30 & 35.12 & 31.16 \\
Merak-7B-v4 (5-shot)             & 41.79 & 45.76 & 42.58 & 35.52 & 42.31 & 42.74 & 42.74 \\
\hline
\multicolumn{8}{l}{\textit{English-medical LLM}} \\
BioMistral-7B (0-shot)           & 27.30 & 32.18 & 30.97 & 23.01 & 27.11 & 33.19 & 52.58 \\
BioMistral-7B (5-shot)           & 42.32 & 32.29 & 43.73 & 37.67 & 35.04 & 53.04 & 43.28 \\
MedGemma-4B-IT (0-shot)          & 43.64 & 40.65 & 41.54 & 42.10 & 44.71 & 51.71 & 37.93 \\
MedGemma-4B-IT (5-shot)          & \textbf{57.58} & 51.25 & 54.41 & 47.96 & \textbf{57.30} & \textbf{66.36} & 50.48 \\
\hline
\end{tabular}
\caption{F1 by disease category on the IndoNutriQA test split. Categories follow the Section~\ref{subsec:domain_coverage} treemap; the test split contains the seven categories shown. Cardiac and renal/urologic are not represented in this test split. Bold marks the strongest model per category.}
\label{tab:disease_breakdown}
\end{table*}

\begin{table*}[!tb]
\centering
\small
\begin{tabular}{l c c c c c c}
\hline
\textbf{Model} & \textbf{Correct} & \textbf{Partial} & \textbf{Wrong span} & \textbf{Term.~mismatch} & \textbf{Hallucination} & \textbf{Noise} \\
\hline
\multicolumn{7}{l}{\textit{Fine-tuned baselines}} \\
Rifky\_IndoBERT\_QA              & 32.0 & 14.0 &  0.0 &  2.0 &  0.0 & 52.0 \\
XLM-R-large                      & 10.0 & 14.0 &  0.0 &  0.0 &  0.0 & 76.0 \\
mDeBERTa-v3-base                 & 12.0 & 16.0 &  0.0 &  0.0 &  0.0 & 72.0 \\
\hline
\multicolumn{7}{l}{\textit{Closed-source LLM (5-shot)}} \\
GPT-4o-mini                      & 30.0 & 70.0 &  0.0 &  0.0 &  0.0 &  0.0 \\
GPT-4o                           & \textbf{44.0} & 52.0 &  0.0 &  0.0 &  0.0 &  4.0 \\
\hline
\multicolumn{7}{l}{\textit{Open-source multilingual LLM (5-shot)}} \\
Llama-3.1-8B-Instruct            & 36.0 & 62.0 &  0.0 &  0.0 &  0.0 &  2.0 \\
Qwen2.5-7B-Instruct              & 30.0 & 70.0 &  0.0 &  0.0 &  0.0 &  0.0 \\
\hline
\multicolumn{7}{l}{\textit{Indonesian-tuned LLM (5-shot)}} \\
Sahabat-AI-Llama3-8B             & 28.0 & 58.0 &  4.0 &  8.0 &  0.0 &  2.0 \\
Merak-7B-v4                      & 26.0 & 68.0 &  6.0 &  0.0 &  0.0 &  0.0 \\
\hline
\multicolumn{7}{l}{\textit{English-medical LLM (5-shot)}} \\
BioMistral-7B                    & 16.0 & 72.0 &  6.0 &  0.0 &  0.0 &  6.0 \\
MedGemma-4B-IT                   & 38.0 & 62.0 &  0.0 &  0.0 &  0.0 &  0.0 \\
\hline
\end{tabular}
\caption{Error taxonomy over a sampled subset of $n = 50$ test examples (seed $42$), classified by GPT-4o-mini. Cells show \% of the $50$ sampled outputs per row; each row sums to $100\%$. Bold marks the strongest model on correct. The Hallucination column is a judge label that human audit shows to under-detect unsupported content (Section~\ref{subsec:human_error}); it is not evidence of absence.}
\label{tab:error_taxonomy}

\end{table*}

\begin{table}[!tbp]
\centering
\small
\resizebox{\columnwidth}{!}{
\begin{tabular}{lrrrr}
\toprule
\multirow{2}{*}{Disease category} & \multicolumn{2}{c}{Source reports} & \multicolumn{2}{c}{QA pairs} \\
\cmidrule(lr){2-3}\cmidrule(lr){4-5}
 & $n$ & \% & $n$ & \% \\
\midrule
Respiratory            & $34$ & $28.33$ & $273$ & $29.97$ \\
Infectious             & $21$ & $17.50$ & $157$ & $17.23$ \\
Gastrointestinal       & $15$ & $12.50$ & $104$ & $11.42$ \\
Nutrition/growth       & $13$ & $10.83$ & $92$  & $10.10$ \\
Neurologic             & $12$ & $10.00$ & $86$  & $9.44$  \\
Cardiac                & $10$ & $8.33$  & $74$  & $8.12$  \\
Hematologic/oncologic  & $10$ & $8.33$  & $85$  & $9.33$  \\
Other                  & $3$  & $2.50$  & $22$  & $2.41$  \\
Renal/urologic         & $1$  & $0.83$  & $11$  & $1.21$  \\
ENT                    & $1$  & $0.83$  & $7$   & $0.77$  \\
\midrule
Total                  & $120$ & $100.00$ & $911$ & $100.00$ \\
\bottomrule
\end{tabular}}
\caption{Disease-category distribution of IndoNutriQA. Each source report is assigned a single dominant category, so the columns partition the corpus; percentages are rounded and may not sum exactly to $100$. The same categorisation backs the stratified results in Section~\ref{subsec:stratified_analysis}.}
\label{tab:disease_distribution}
\end{table}

\subsection*{Error Taxonomy}

Table~\ref{tab:error_taxonomy} classifies each model's outputs on a sampled subset of $50$ test examples into six categories (correct, partial, wrong span, terminology mismatch, hallucination, and noise), assigned by an LLM judge (GPT-4o-mini). This qualitative view complements the aggregate metrics in Section~\ref{subsec:model_performance} by revealing how models fail rather than only how much. Two patterns stand out. First, the fine-tuned extractive baselines are dominated by noise ($52$--$76\%$), reflecting empty or malformed span predictions, whereas the $5$-shot LLMs convert almost all of these into partial answers that capture some but not all of the reference content. Second, and most importantly, the human error analysis of Section~\ref{subsec:human_error} shows that the judge's hallucination label cannot be relied on: raters reading the same class of output flagged unsupported additions in $10\%$ of audited units, including two that assert a clinical trend the patient record does not support and that the judge scored $4$ on factuality. We therefore draw no conclusion about fabrication rates; under-answering is the dominant failure on both the judge's and the raters' accounts.

\section{Demographic, Anthropometric, and Syntactic-Divergence Figures}
\label{appx:demographics}

\begin{figure}[!tbp]
    \centering
    \includegraphics[width=0.9\columnwidth]{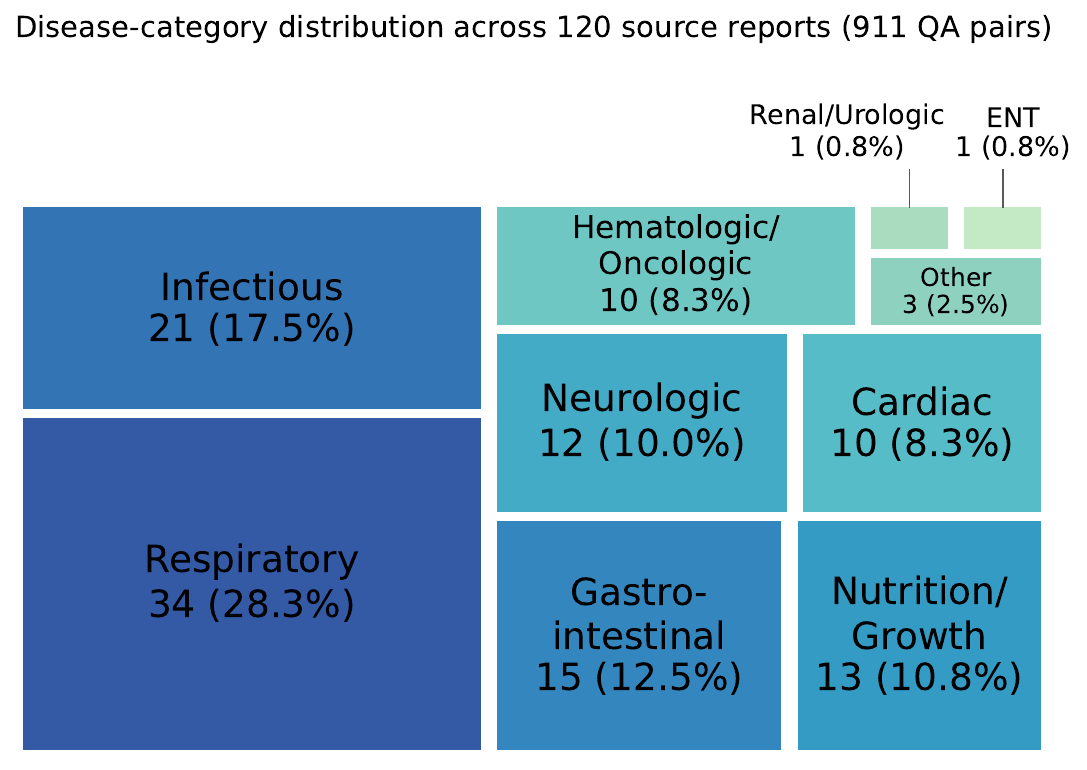}
    \caption{Distribution of disease categories in IndoNutriQA.}
    \label{fig:treemap-medical-diagnoses}
\end{figure}

\begin{figure}[!tbp]
  \centering
  \subfloat[Age (months)\label{fig:age}]{\includegraphics[width=.8\columnwidth]{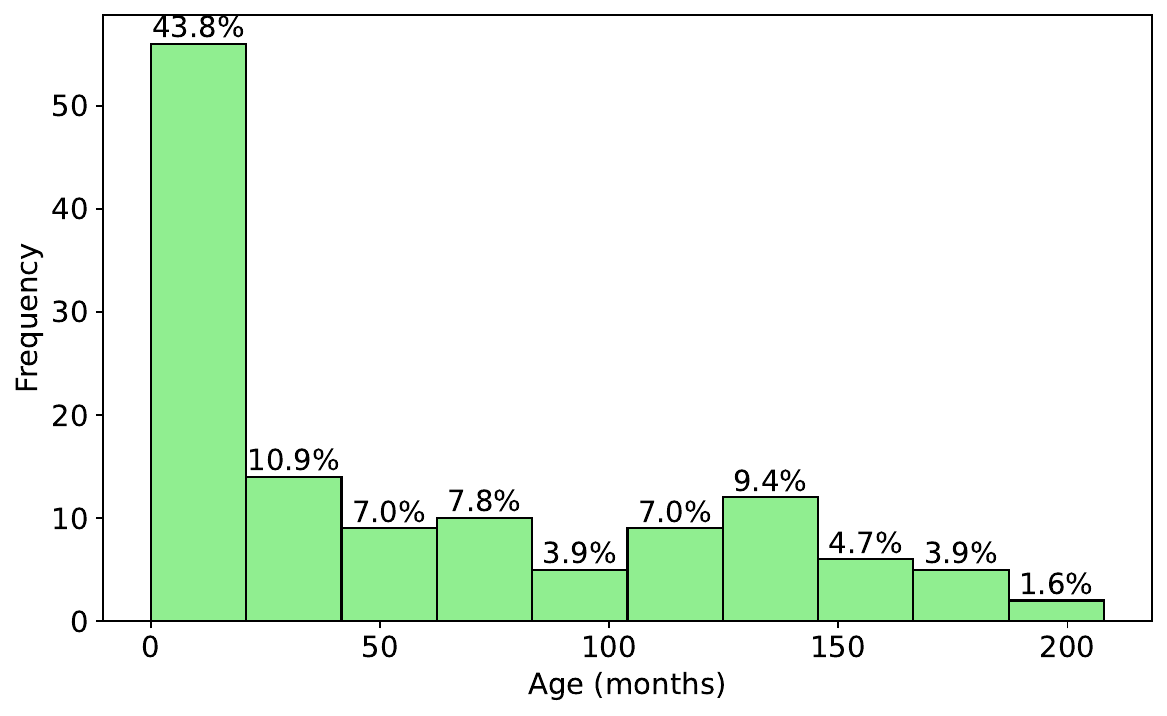}}\\
  \subfloat[Sex\label{fig:sex}]{\includegraphics[width=.8\columnwidth]{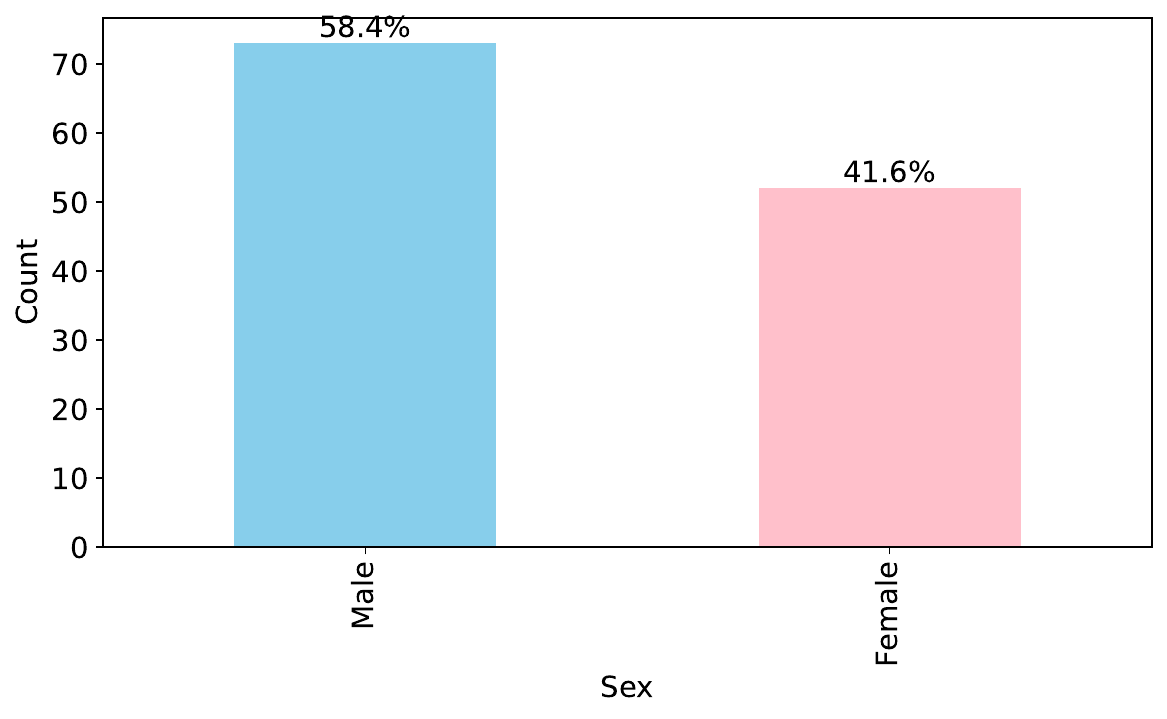}}\\
  \subfloat[Weight vs height\label{fig:weight_height}]{\includegraphics[width=.8\columnwidth]{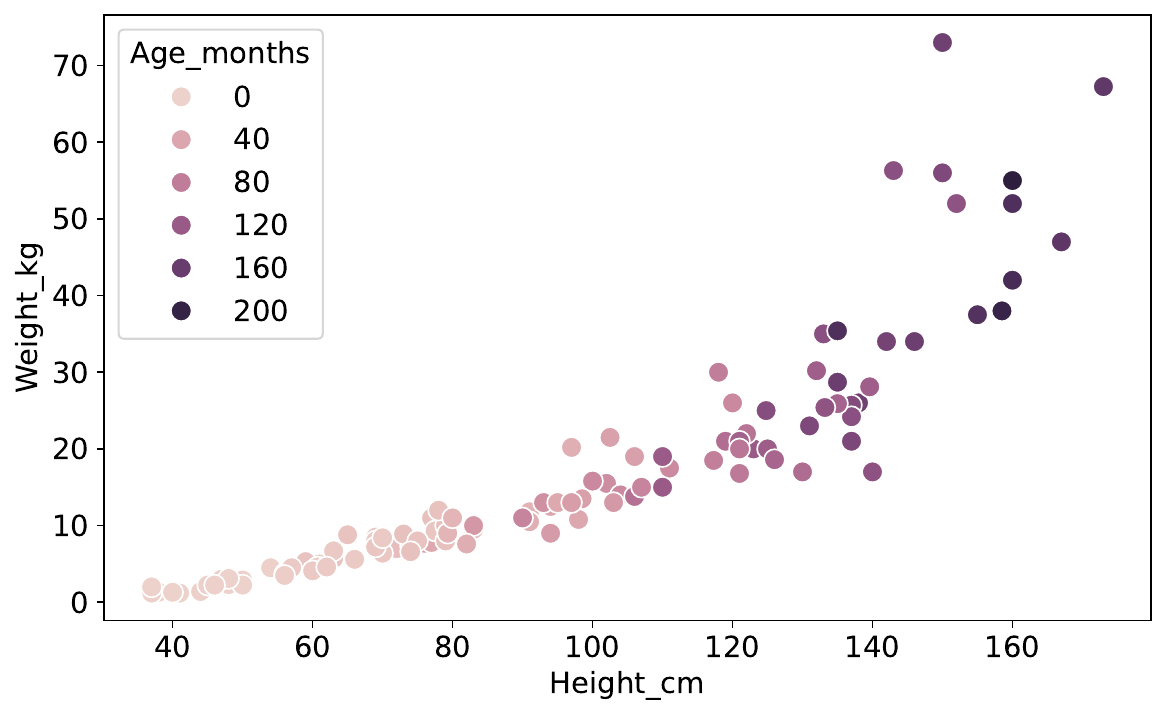}}\\
  \subfloat[MUAC (LILA)\label{fig:lila}]{\includegraphics[width=.8\columnwidth]{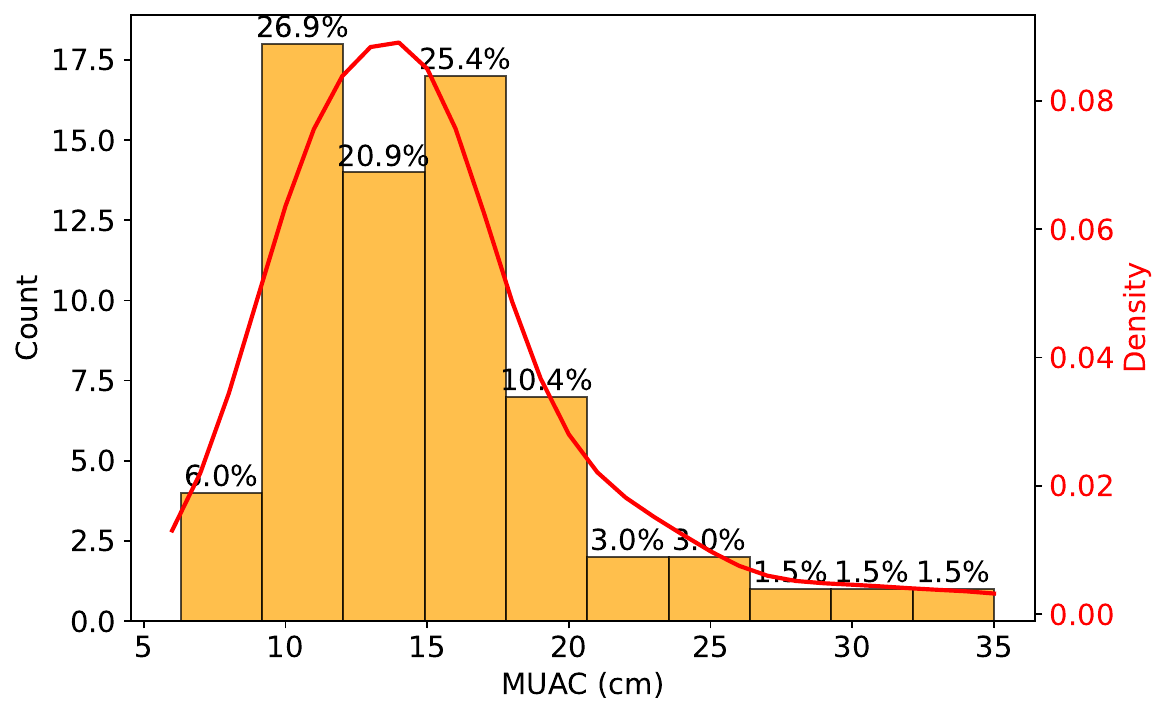}}\\
  \subfloat[Syntactic divergence\label{fig:syntactic}]{\includegraphics[width=.8\columnwidth]{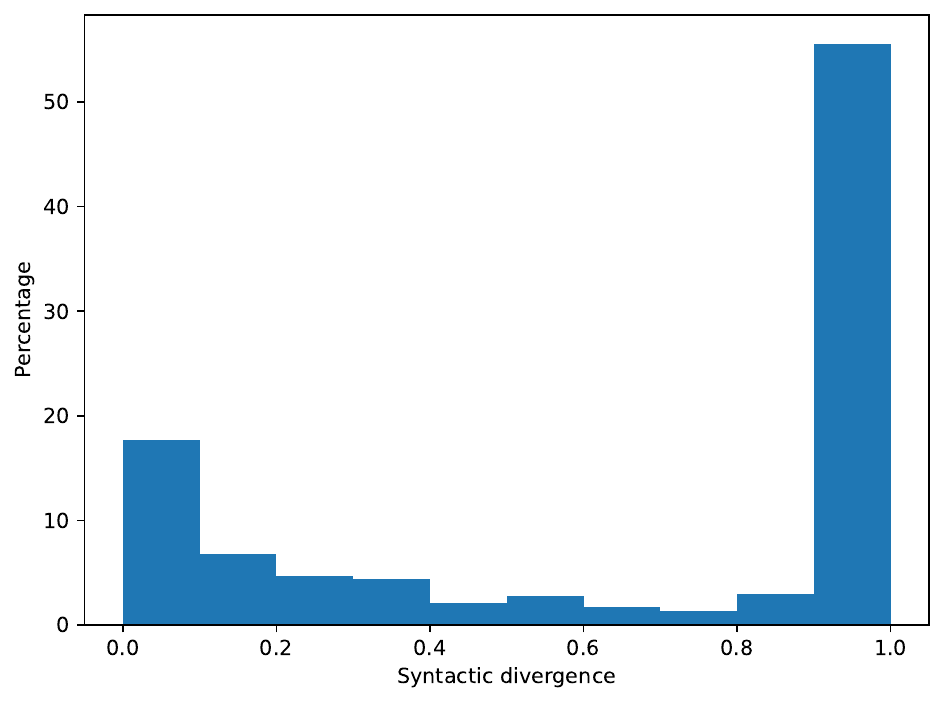}}
  \caption{Demographic and anthropometric characteristics of IndoNutriQA cases, and distribution of token-level syntactic divergence between model predictions and reference answers.}
  \label{fig:character}
\end{figure}

Table~\ref{tab:disease_distribution} gives the number and share of source reports and QA pairs for each disease category. Figure~\ref{fig:treemap-medical-diagnoses} presents the full distribution of disease categories in IndoNutriQA. Figure~\ref{fig:character} shows the age in months (panel~(a)), sex (panel~(b)), weight-vs-height colored by age (panel~(c)), and MUAC/LILA (panel~(d)) distributions of patient cases, alongside the bimodal distribution of token-level normalized edit distance between model predictions and reference answers (panel~(e)).

\section{Standardized Questions and Example Instance}
\label{sec:appendix}

\subsection*{Standardized Clinical Nutrition Questions}
\begin{enumerate}
    \item Apa metode skrining status gizi yang direkomendasikan? \\
    \textit{What nutritional screening method is recommended?}
    \item Bagaimana status gizi pada pasien ini? \\
    \textit{What is the nutritional status of this patient?}
    \item Bagaimana data biokimia/parameter lab pada pasien (yang abnormal)? \\
    \textit{What are the abnormal biochemical or laboratory parameters of this patient?}
    \item Bagaimana menyusun diagnosis gizi dengan format PES (Problem, Etiology, Sign and Symptom) pada pasien ini? \\
    \textit{How is the nutrition diagnosis constructed in PES (Problem, Etiology, Sign and Symptom) format for this patient?}
    \item Apa tujuan diet pada pasien ini? \\
    \textit{What are the dietary goals for this patient?}
    \item Bagaimana prinsip diet yang tepat sesuai dengan kondisi pasien ini? \\
    \textit{What are the appropriate dietary principles according to this patient’s condition?}
    \item Bagaimana syarat diet yang harus dipenuhi dalam pengaturan diet pasien ini? \\
    \textit{What dietary requirements must be fulfilled in this patient’s diet management?}
    \item Apa pedoman diet yang digunakan dalam perencanaan diet pasien ini? \\
    \textit{What dietary guidelines are used in planning the patient’s diet?}
\end{enumerate}

\section{Example Question-Answer Instance}
\label{appendix:qa_examples}

This subsection presents representative example of question-answer instance from the IndoNutriQA dataset. The example includes the original Indonesian text accompanied by an English translation to facilitate understanding for non-Indonesian readers. All personally identifiable information has been anonymized.

\paragraph{Clinical Context (Excerpt).}
\begin{quote}
\small
\textbf{Report Title:} Laporan Praktik Kerja Profesi Program Studi Profesi Dietisien Rotasi Gizi Klinik Sub Rotasi Rawat Inap (Stase PICU/NICU)

\textit{Professional Clinical Internship Report for the Dietitian Profession Program Clinical Nutrition Rotation – Inpatient Sub-Rotation (PICU/NICU Stage)}

\medskip \textbf{Anthropometric Assessment:} Jenis kelamin: Laki-laki. Usia: 1 tahun 1 bulan. Panjang badan (PB): 65 cm. BBI/U menurut PB: 8,8 kg.

\textit{Sex: Male. Age: 1 year 1 month. Body length (PB): 65 cm. Ideal body weight for length: 8.8 kg.}

\medskip \textbf{Medical Diagnosis:} Severe Obstructive Hydrocephalus post repair VP shunt + Labiopalatoschisis + Cerebral Palsy (CP) + Pneumonia.

\medskip \textbf{Context:} Skrining gizi bertujuan untuk mengetahui apakah pasien berisiko malnutrisi atau tidak. Beberapa alat skrining gizi yang dapat digunakan pada pasien anak-anak adalah Strong Kids, PYMS, SGNA, dan SPNRS.

\textit{Nutritional screening aims to determine whether the patient is at risk of malnutrition. Several screening tools that can be used for pediatric patients include Strong Kids, PYMS, SGNA, and SPNRS.}
\end{quote}

\paragraph{Question.} Apa metode skrining status gizi yang direkomendasikan?

\textit{What nutritional screening method is recommended?}

\paragraph{Answer (Extractive).} Beberapa alat skrining gizi yang dapat digunakan pada pasien anak-anak adalah Strong Kids, PYMS, SGNA, dan SPNRS.

\textit{Several nutritional screening tools that can be used for pediatric patients are Strong Kids, PYMS, SGNA, and SPNRS.}

\end{document}